\documentclass[lettersize,journal]{IEEEtran}

\usepackage{amsmath,amsfonts}
\usepackage{algorithmic}
\usepackage{array}
\usepackage[caption=false,font=normalsize,labelfont=sf,textfont=sf]{subfig}
\usepackage{textcomp}
\usepackage{stfloats}
\usepackage{url}
\usepackage{verbatim}
\usepackage{graphicx}
\usepackage{cite}

\usepackage{xcolor}
\usepackage{balance}
\usepackage[breaklinks=true]{hyperref}
\usepackage{multirow}
\usepackage{booktabs}
\usepackage{graphicx}
\usepackage{siunitx}
\usepackage[super]{nth}
\usepackage{transparent}
\usepackage{flushend}
\usepackage{lipsum}
\usepackage{tikz}

\def\BibTeX{{\rm B\kern-.05em{\sc i\kern-.025em b}\kern-.08em
    T\kern-.1667em\lower.7ex\hbox{E}\kern-.125emX}}

\newcommand{\mycomment}[1]{}

\graphicspath{{figures/APs}{figures/bev}}

\newcommand\copyrighttext{%
    \footnotesize \textcopyright 2023 IEEE.  Personal use of this material is permitted.  Permission from IEEE must be obtained for all other uses, in any current or future media, including reprinting/republishing this material for advertising or promotional purposes, creating new collective works, for resale or redistribution to servers or lists, or reuse of any copyrighted component of this work in other works. DOI: \href{https://doi.org/10.1109/TIV.2023.3251650}{10.1109/TIV.2023.3251650}
}
\newcommand\copyrightnotice{%
    \begin{tikzpicture}[remember picture,overlay]
    \node[anchor=south,yshift=0pt, xshift=10pt] at (current page.south) {\fbox{\parbox{\dimexpr\textwidth-\fboxsep-\fboxrule\relax}{\copyrighttext}}};
    \end{tikzpicture}%
}

\begin{document}
\title{Quantifying the LiDAR Sim-to-Real Domain Shift: A Detailed Investigation Using Object Detectors and Analyzing Point Clouds at Target-Level}
\author{Sebastian~Huch, Luca~Scalerandi, Esteban~Rivera, Markus~Lienkamp
\thanks{Manuscript received Month XX, XXXX; revised Month XX, XXXX.}%
\thanks{Sebastian~Huch, Esteban~Rivera, and Markus~Lienkamp are with the Institute of Automotive Technology, School of Engineering and Design, Technical University of Munich, Boltzmannstraße 15, 85748 Garching, Germany (e-mail: sebastian.huch@tum.de, esteban.rivera@tum.de, lienkamp@tum.de)}
\thanks{Luca~Scalerandi is with the Department of Informatics, Technical University of Munich, Boltzmannstraße 3, 85748 Garching, Germany (e-mail: luca.scalerandi@tum.de)}}

\markboth{IEEE TRANSACTIONS ON INTELLIGENT VEHICLES,~Vol.~XX, No.~XX, Month~XXXX}%
{Huch \MakeLowercase{\textit{et al.}}: Quantifying the LiDAR Sim-to-Real Domain Shift: A Detailed Investigation Using Object Detectors and Analyzing Point Clouds on Target-Level}

\maketitle
\copyrightnotice

\begin{abstract}
LiDAR object detection algorithms based on neural networks for autonomous driving require large amounts of data for training, validation, and testing.
As real-world data collection and labeling are time-consuming and expensive, simulation-based synthetic data generation is a viable alternative.
However, using simulated data for the training of neural networks leads to a domain shift of training and testing data due to differences in scenes, scenarios, and distributions.
In this work, we quantify the sim-to-real domain shift by means of LiDAR object detectors trained with a new scenario-identical real-world and simulated dataset.
In addition, we answer the questions of how well the simulated data resembles the real-world data and how well object detectors trained on simulated data perform on real-world data.
Further, we analyze point clouds at the target-level by comparing real-world and simulated point clouds within the 3D bounding boxes of the targets.
Our experiments show that a significant sim-to-real domain shift exists even for our scenario-identical datasets.
This domain shift amounts to an average precision reduction of around \SI{14}{\percent} for object detectors trained with simulated data.
Additional experiments reveal that this domain shift can be lowered by introducing a simple noise model in simulation.
We further show that a simple downsampling method to model real-world physics does not influence the performance of the object detectors.
\end{abstract}


\section{Introduction}
\IEEEPARstart{A}{utonomous} vehicles (AVs) have the potential to increase road safety and reduce emissions \cite{Costley2021LowResearch}.
Therefore, AV research and development has made great strides forward in recent years to such an extent that the first autonomous shuttles with limited operational design domain (ODD) are driving on public roads.
Nevertheless, those are hand-tailored examples of specific situations which can not be easily generalized.
In order to extend those ODDs, a reliable software pipeline consisting of sequential (interconnected) modules for perception, prediction, planning, and control is required.
In such a pipeline, the perception module estimates the state of the AV with respect to its environment, through sensors like camera, LiDAR, and RADAR, and provides the relevant information needed to perform the driving task to the subsequent modules.
Part of the perception module is object detection algorithms, which use the sensor readings as an input and the output is a list of objects.
These lists contain the object's position, shape, and orientation, and represent the different traffic participants present in the environment.

These object detection algorithms are highly dependent on machine learning to extract features from raw sensor data.
For example, deep neural networks such as YOLOv3 \cite{Redmon2018YOLOv3:Improvement} or PV-RCNN \cite{Shi2019PV-RCNN:Detection} are used to detect objects in camera images or LiDAR point clouds, respectively.

\begin{figure}[t!]
	\centering
 	\subfloat[\label{fig:full_pcd_bev_sim}]
  {\selectfont\def\svgwidth{1.0\columnwidth}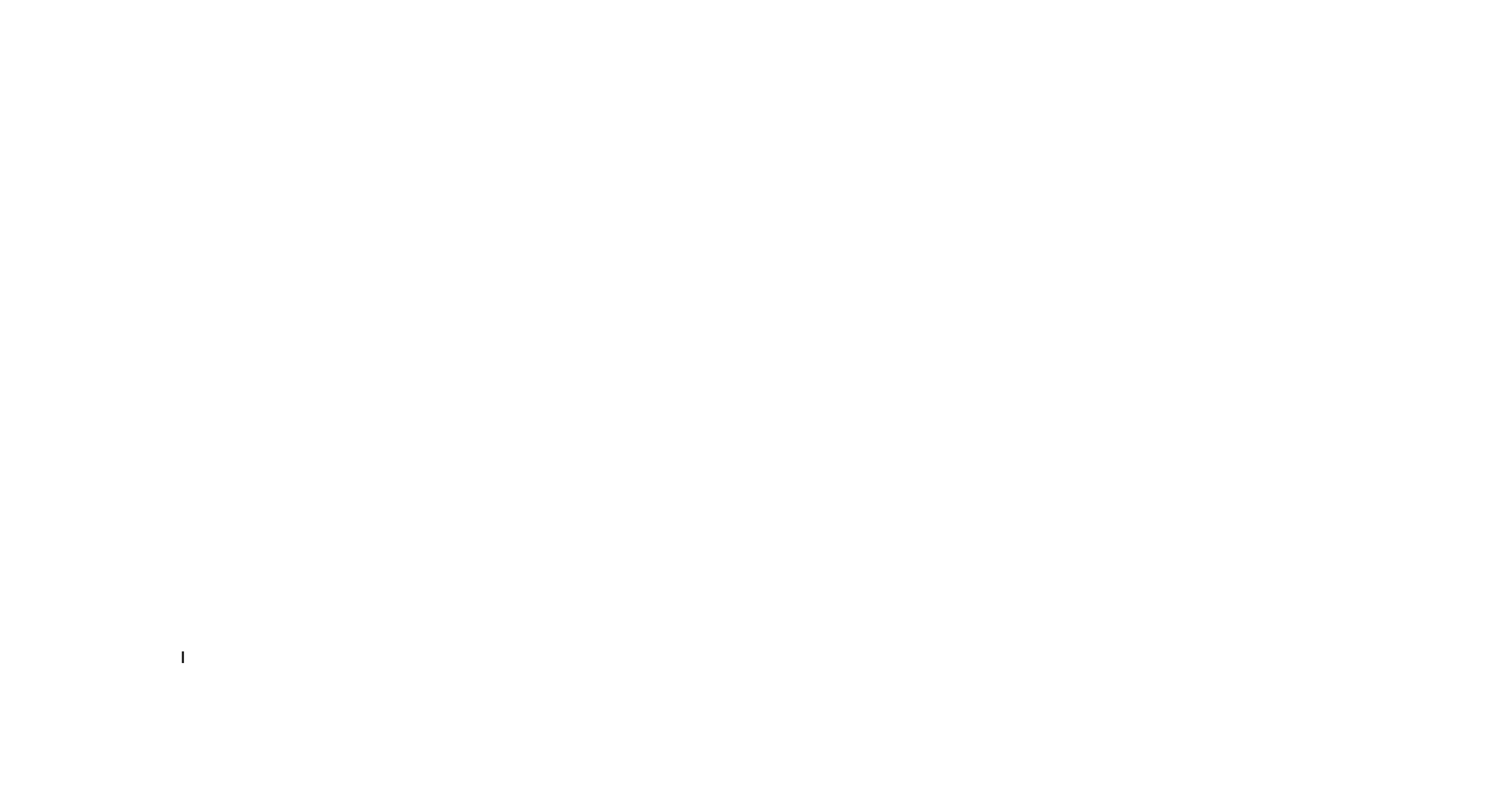}
        \hfill
	\subfloat[\label{fig:full_pcd_bev_real}]
 {\selectfont\def\svgwidth{1.0\columnwidth}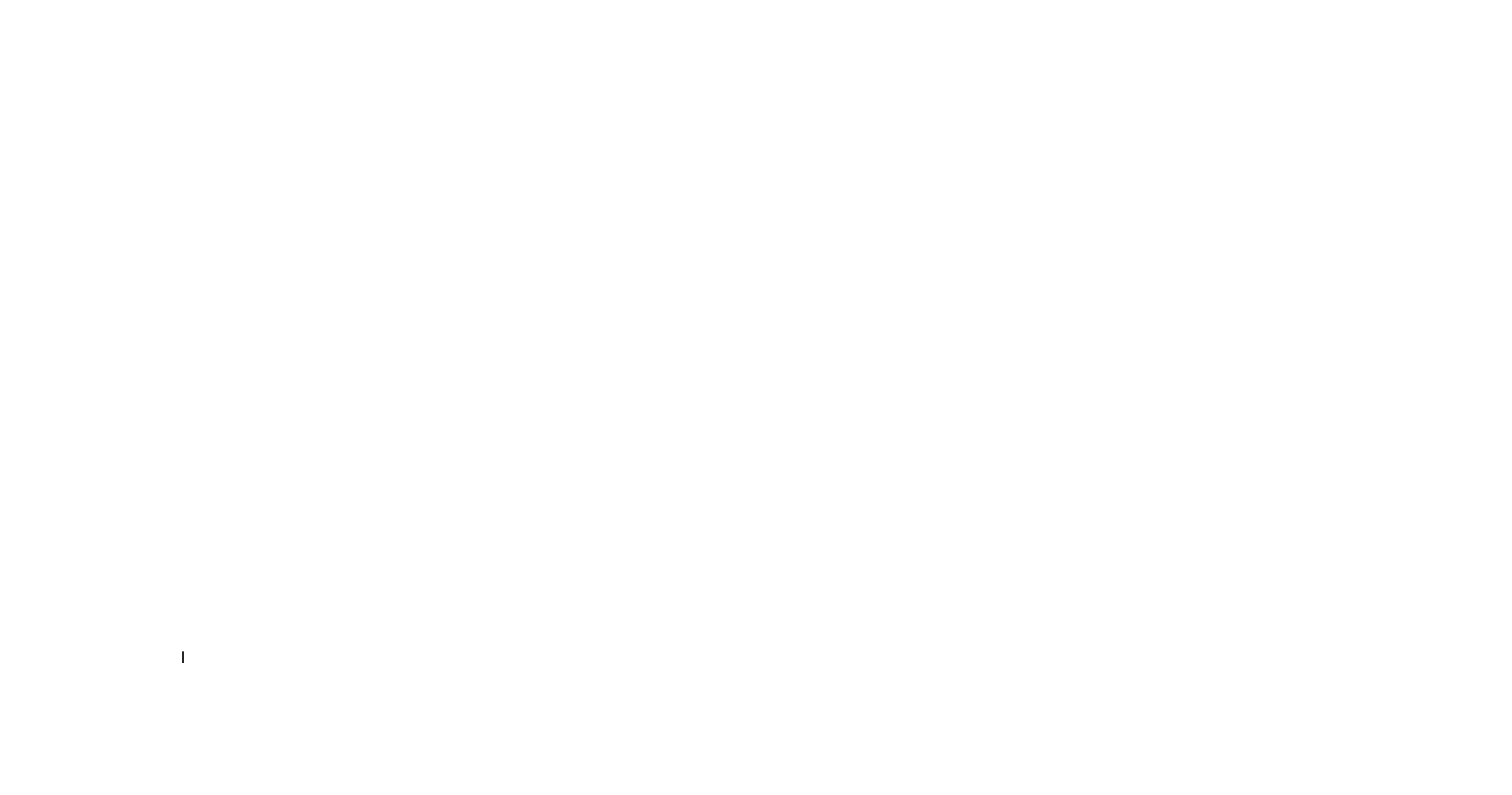}
	\caption{Bird's-eye-view of point clouds of our (a) \textit{sim} and (b) \textit{real} datasets. The driving direction is to the right.}\label{fig:full_pcd_bev}
\end{figure}
Most of the approaches to train those networks are based on supervised learning, meaning they need labeled data.
Depending on the data domain, labeling individual frames or point clouds can be very costly in terms of time, effort, and money.
Specifically, for a real-world AV dataset, a vehicle equipped with cameras and LiDAR sensors records sensor data from specific scenarios.
Afterward, the captured dataset is manually labeled frame by frame.
Additionally, such a dataset can become outdated quickly due to constant upgrades to the AV's sensors, e.g., higher resolution of a new LiDAR sensor.

An alternative to labeled real-world data is synthetic data, which can be categorized into augmented real-world data or simulated data.
The former can be generated, e.g., by extracting objects from an already labeled real-world dataset and placing these objects into other frames of that dataset.
This method is known as domain randomization and is performed to improve an object detector's performance, for example, for camera object detection \cite{Tremblay2018TrainingRandomization}.
Although the results of domain randomization are promising, the dataset is still limited to the scenarios captured during the initial dataset generation.

The second form of synthetic data is a dataset generated in a simulation environment with virtual sensor models.
These simulations can model a photo-realistic 3D environment and include a physics simulation for static and dynamic actors, such as vehicles and pedestrians.
Several open-source simulations specially designed for autonomous driving are available, e.g., CARLA \cite{Dosovitskiy2017CARLA:Simulator}.
The main advantages of datasets generated in the simulation are the unlimited data generation, including automatic pointwise labels, the option to vary environment conditions and sensor configurations, and rapid scenario construction, which allows the simulation of potentially dangerous edge cases.

The effectiveness of a dataset is not only defined by the quantity of data but also by the quality of the data in terms of realism, distribution, and diversity \cite{Nowruzi2019HowData}.
Although neural networks can be trained with large simulated datasets, this does not per se guarantee good performance in a real-world application.
The success of training and testing on different domains, for example, simulated and real-world, as in Fig.~\ref{fig:full_pcd_bev}, is based on the assumption that both domains share the same feature space and distribution \cite{Triess2021APerception}.
However, in real-world applications, this assumption is not satisfied. 
Existing simulation environments do not generate sensor data identical to real-world data due to the fact that simulated sensor models do not model both real-world physics and sensor characteristics accurately.
Therefore, the performance of object detection algorithms trained with simulated data is affected by this phenomenon, which is called domain shift.
More generally, the term domain shift, also called domain gap, refers to the difference between two domains, as observed through the available data.
The analysis of this domain shift and its quantification in the scope of LiDAR sensors are the main content of this paper.

More specifically, our main contributions are as follows:
\begin{itemize}
	\item We propose a method to generate scenario-identical datasets which can be used to quantify the sim-to-real domain shift using 3D LiDAR object detectors.
	\item We quantify the domain shift of simulated and real-world datasets and further break down the results by exploiting a point cloud target-level analysis.
	\item We show that simple modifications to the virtual sensor model, such as noise, can increase the object detector's performance, and thus lower the sim-to-real domain shift.
	\item We provide a labeled distribution-aligned (scenario-identical) LiDAR point cloud dataset with simulated and real-world data for the evaluation of future domain adaptation approaches: \href{https://github.com/TUMFTM/Sim2RealDistributionAlignedDataset}{\nolinkurl{https://github.com/TUMFTM/Sim2RealDistributionAlignedDataset}}.
\end{itemize}
This work is organized as follows:
Sec.~\ref{sec:related_work} discusses the approaches of similar works in this research area.
In Sec.~\ref{sec:method}, we first explain the data extraction for our real-world dataset and the data generation in simulation.
This is followed by introducing domain shift quantification, including neural networks, metrics for evaluation, and point cloud target-level analysis.
The results of the experiments and their discussion can be found in Sec.~\ref{sec:results} and Sec.~\ref{sec:discussion}, respectively.
We summarize our work in Sec.~\ref{sec:conclusion}.

\section{Related Work}
\label{sec:related_work}

The performance discrepancy between the unequal source and target domains has been studied for years \cite{Ben-David2010ADomains}.
Even minimal changes in the source domain, like a camera parameter change, can have big effects on a model's performance \cite{Dodge2016UnderstandingNetworks}.
Neural networks, even though trained on similar samples from one domain, frequently underperform with reference to test data from other domains \cite{Ben-David2010ADomains}, \cite{Wang2020TrainGeneralize}.

Several works cover the quantification of the domain shift, but many of them refer to camera images instead of LiDAR point clouds.

Adam et al. \cite{Adam2019RobustnessDriving} and Nowruzi et al. \cite{Nowruzi2019HowData} train multiple camera 2D object detectors on simulated data and test the networks on a mixed dataset containing real-world and simulated images.
They calculate standard object detection metrics, such as mean average precision (mAP), to quantify the domain shift.
Both works conclude that a domain shift is noticeable by a difference in the mAP of networks trained on simulated or real-world data.
\cite{Nowruzi2019HowData} further investigate the potential of training with mixed datasets and fine-tuning networks on the target (real-world) dataset.
Their studies show that fine-tuning performs better than mixed data training and that fine-tuning can lower but not eliminate the domain shift.
The performance of mixed data training is also the research topic of Seib et al. \cite{Seib2020MixingApproaches} and Burdorf et al. \cite{Burdorf2022ReducingData}.
Their results indicate that synthetic data can replace real-world data to an extent, but proportionately more synthetic than real-world data is required. Moreover, synthetic data can be successfully used for network pre-training leading to better performance compared to training on real-world data only \cite{Gaidon2016VirtualAnalysis}\cite{Cabon2020Virtual2}.

Similar research to quantify the domain shift has also been conducted for LiDAR point clouds with 3D object detectors.
Dworak et al. \cite{Dworak2019PerformanceSimulator} train three object detectors with data generated in the simulation environment CARLA and evaluate their performance on the real-world KITTI dataset \cite{Geiger2013VisionDataset}.
Although networks achieve an mAP of up to 87~\% when trained and evaluated on CARLA (``sim-to-sim''), the best network only reaches an mAP of 19~\% if trained on CARLA and tested on KITTI (``sim-to-real'').
The authors also experimented with fine-tuning and mixed-data training.
The results of these methods follow the findings of \cite{Nowruzi2019HowData} for camera 2D object detectors.

To evaluate their methods for synthetic LiDAR data generation, Fang et al. \cite{Fang2018AugmentedDriving} and Manivasagam et al. \cite{Manivasagam2020LiDARsim:World} compared their generated data with the CARLA and KITTI datasets by training object detectors and evaluating on KITTI.
Both works highlight the sim-to-real domain shift between the CARLA and KITTI datasets.

Yue et al. \cite{Yue2018ADriving} and Spiegel et al. \cite{Spiegel2021UsingClouds} conduct similar LiDAR sim-to-real comparison experiments but focus on semantic segmentation instead of object detection.
However, the sim-to-real domain shift is also measurable with networks designed for different tasks, such as segmentation.

The works of Tsai et al. \cite{Tsai2022SeeAdaptation} and Wang et al. \cite{Wang2020TrainGeneralize} investigate the real-to-real domain shift, which occurs when a neural network is trained and tested on different real-world datasets.
While \cite{Tsai2022SeeAdaptation} focus on the difference in the scan pattern of the LiDAR sensors used in the KITTI, nuScenes \cite{Caesar2019NuScenes:Driving}, and Waymo \cite{Sun2020ScalabilityDataset} datasets, \cite{Wang2020TrainGeneralize} point out the statistical differences in vehicle shapes and sizes in the datasets collected in different countries.
\cite{Wang2020TrainGeneralize} also suggest a domain adaptation approach using statistical normalization to improve cross-dataset performance.

The existence of a domain shift leads to the study of domain adaptation which is concerned with creating models, adapting data, and applying other techniques to allow models to generalize well to a target domain even though they have been trained on a different source data distribution \cite{Ben-David2010ADomains}.
Several works propose domain adaptation methods to minimize camera domain shift, such as domain randomization \cite{Tremblay2018TrainingRandomization}\cite{Prakash2018StructuredData}\cite{Yue2019DomainData}, domain augmentation \cite{AbuAlhaija2017AugmentedScenes}, or generative adversarial networks \cite{Lin2020GENERATINGNETWORKS}.
Similar efforts also exist that cover the LiDAR domain shift, but they directly target synthetic data generation \cite{Yue2018ADriving}\cite{Fang2018AugmentedDriving} instead of modifying existing datasets.
All the mentioned domain adaptation works benchmark the effectiveness of their method by training object detectors with the synthetic (source), the adapted synthetic, and the real-world (target) datasets and evaluate the performance on the real-world dataset.

Ljungqvist et al. \cite{Ljungqvist2022ObjectData} compare 2D object detectors trained with synthetic and real-world image data with a different method.
Instead of calculating metrics based on the network's final outputs, i.e., the 3D bounding boxes of the predicted objects, the authors compare the similarity of the outputs of each network layer.
For each layer, they calculate the similarity index \textit{linear centered kernel alignment} (CKA) \cite{Kornblith2019SimilarityRevisited} and conduct a layer-wise comparison of networks trained with synthetic and real-world data.
The analysis shows a high similarity in the early network layers and a relatively low similarity in the network detection head.

Triess et al. \cite{Triess2022AClouds} define a metric to quantify the realism of generated LiDAR point clouds.
This metric is based on an adversarial learning technique and can be applied to unseen data.
They prove the effectiveness of their quantitative metric by evaluating semantic segmentation networks.

We address the limitations of the related work.
Similar works quantified the domain shift based on synthetic and real-world datasets with different scenes, scenarios, and distributions, e.g., by using a dataset recorded in the simulation environment CARLA and the real-world Waymo dataset.
This approach does not allow one to conclude whether the domain shift originates from an unrealistic sensor model in simulation, from the distribution shift, or a combination of both.
We aim to investigate the domain shift of synthetic and real-world point clouds using datasets with identical scenarios and distributions. which we will explain in the following sections.

\begin{figure}[!t]
  \centering
  \subfloat[]{\includegraphics[width=0.8\linewidth]{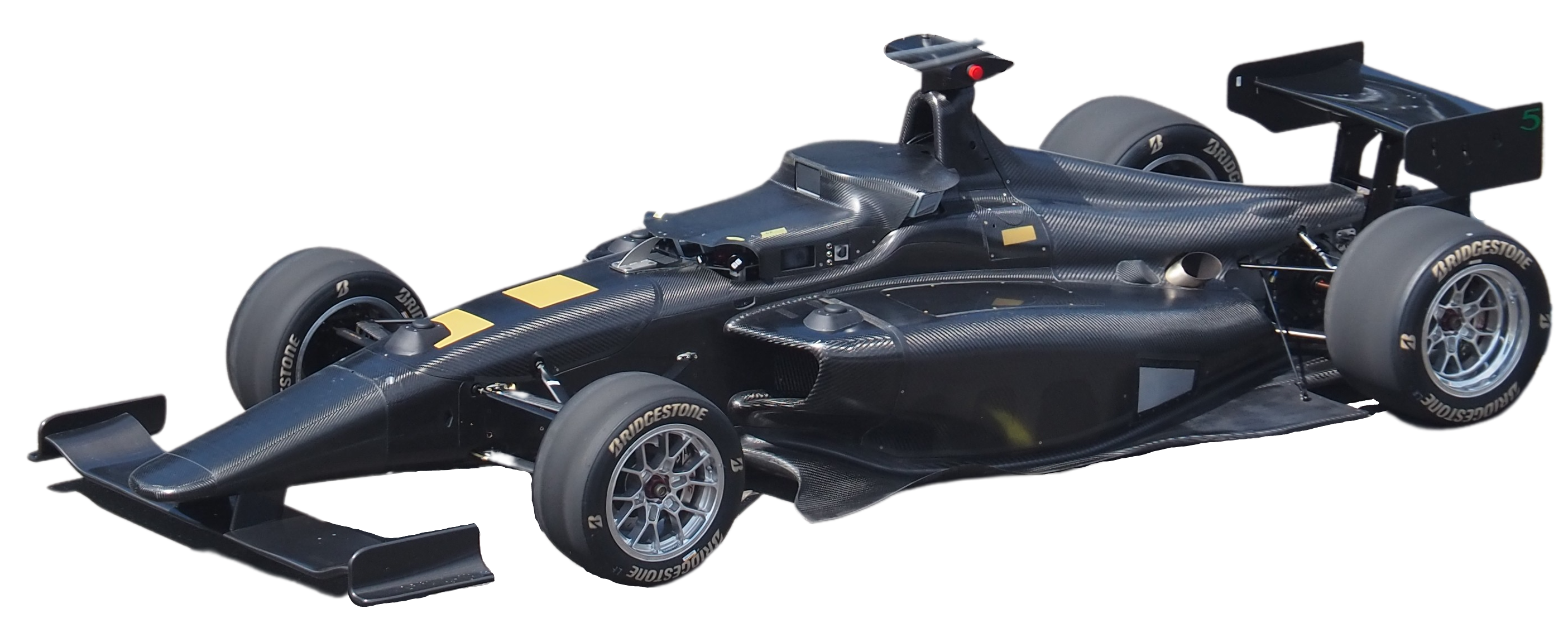}%
  \label{fig:car_real}}
  \vfil
  \subfloat[]{\includegraphics[width=0.8\linewidth]{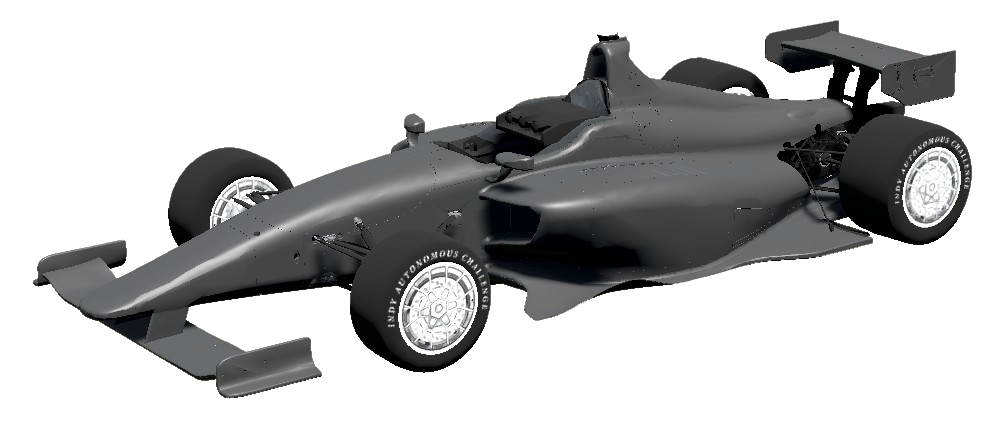}%
  \label{fig:car_sim}}
  \caption{(a) Picture of the real-world Dallara AV-21 race car and (b) a 3D model of the Dallara AV-21 in the customized simulation environment Unity used to generate the simulated dataset.}
  \label{fig:3d_models}
\end{figure}

\section{Methodology}
\label{sec:method}
In this section, we present our method for the quantification and analysis of the sim-to-real domain shift.
The method consists of three consecutive steps, namely dataset generation, object detection algorithms selection and training, and performance evaluation.
First, we capture real-world data, develop a pipeline to automatically label the data, and generate a simulated dataset based on real-world data.
To validate the generated datasets, we perform a high-level comparison based on statistical parameters; this is described in Sec.~\ref{sec:method_dataset_generation} and \ref{sec:method_stat_data_comp}.
The second step of our method compromises the selection of neural networks for object detection.
Together with the explanation of the configuration and training parameters of the networks, this is described in Sec.~\ref{sec:method_networks}.
The last step of our method is the evaluation of the selected networks trained with the generated datasets.
We present KPIs used for domain shift quantification in Sec.~\ref{sec:method_kpis}.

\subsection{Dataset Generation}
\label{sec:method_dataset_generation}

To investigate the nature of the sim-to-real domain shift, a novel dataset is used, as none of the existing datasets fulfills the requirement of being distribution-aligned.
This dataset consists of two subsets representing the same environment, agents, and scenarios.
These subsets include the \textit{real} dataset and a \textit{sim} dataset derived from the real-world counterpart.
Not only will the  same scene and vehicles be used as in the \textit{real} dataset, but also all driving scenarios are replayed in the simulated environment so that the discrepancy between the \textit{real} and \textit{sim} datasets is limited to a minimum.
Thus, the domain shift can be isolated leaving out possible external influences which could influence its quantification.

All real-world measurement data was captured during the Indy Autonomous Challenge (IAC) in Las Vegas in 2022.
The IAC was the world's first head-to-head autonomous race car competition between international universities.
The race cars used were the Dallara AV-21 (Fig.~\ref{fig:car_real}), a modified Dallara IL-15, equipped with various sensors and programmed to drive autonomously.
During the single and multi-vehicle races along the oval race track, large amounts of sensor data were collected.
Each race car is equipped with three identical LiDAR sensors, each with a horizontal field of view (FoV) of \SI{120}{\degree}, to cover a combined FoV of \SI{360}{\degree} at \SI{20}{\hertz} up to a range of \SI{250}{\meter} (at \SI{10}{\percent} surface reflection).
LiDAR point clouds with pointwise $x$-, $y$-, $z$-, and $intensity$-values, as well as the GPS trajectories of all vehicles, were among the data collected during the training sessions and the actual competition.

Capturing the GPS coordinates of every vehicle is not only beneficial for auto-labeling the real-world point clouds but also allows precise re-simulation and, thus, the generation of the simulated dataset.
The GPS data was captured at a rate of \SI{20}{\hertz} and included the position, orientation, and velocity of each vehicle.
We match the GPS coordinates of all vehicles with the point clouds in each time step to create the labels w.r.t. the ego vehicle's local coordinate system.
In the second step, these labels are refined by calculating the point distribution within and around each 3D bounding box and shifting boxes based on the point distribution to mitigate potential labeling errors. 

The real-world data was used to create the scenario-identical synthetic dataset using a 3D simulator.
The simulator is based on Unity and includes 3D models of Dallara AV-21 (Fig.~\ref{fig:car_sim}) vehicles, the 3D environment of the race track, and a custom LiDAR sensor model.
This sensor model was configured to match the characteristics of the real-world LiDAR sensor.
The sensor model is capable of calculating pointwise $intensity$-values based on the ray incidence angle and target material.
However, the intensities are not validated and therefore not used in the following domain shift quantification and we focus on $x$, $y$, and $z$.

Each dataset consists of $32{,}951$ labeled point clouds from three runs on the race track with several laps, each with speeds of up to \SI{70}{\meter\per\second}.
As the following frames show high similarity, we only select every \nth{5} point cloud for neural network training, validation, and testing.
The final datasets consist of $6{,}000$ individual point clouds divided into a training, validation, and test set in the ratios $\frac{4}{6}$, $\frac{1}{6}$, and $\frac{1}{6}$, respectively.
In the following, we refer to the recorded real-world and generated simulation datasets as \textit{real} and \textit{sim} datasets, respectively.

\subsection{Statistical Dataset Comparison}
\label{sec:method_stat_data_comp}

To investigate the domain shift between the \textit{real} and \textit{sim} data, both datasets should be compared.
All comparisons are made using the data that are used to train the models.

To analyze any discrepancies, we iteratively compare the \textit{real} and \textit{sim} data pairs. 
Therefore, we use the training data loader and extract the data just before the model starts training. 
This way ensures that all analyzed samples correspond to what the model will encounter later during training. 
The fact that the \textit{real} and \textit{sim} datasets are scenario-identical allows the loading of pairs of samples showing the same scene on the track.
All comparisons are based on these corresponding samples.

All valid samples, i.e., the ground truth bounding boxes in the observable range, from one dataset are loaded and matched to the samples of the other dataset based on their time-stamp-correspondent ID.
After the matching, the samples are sorted according to their distance from the ego vehicle. 
We use three different ranges:
\begin{itemize}
    \item close-range $r_1 = \left[\SI{0.0}{\meter}, \SI{33.3}{\meter}\right[$, 
    \item mid-range $r_2 = \left[\SI{33.3}{\meter}, \SI{66.6}{\meter}\right[$, and
    \item long-range $r_3 = \left[\SI{66.6}{\meter}, \SI{100.0}{\meter}\right]$.
\end{itemize}
This distinction allows us to better understand the effects of range on performance metrics. 
Each data sample consists of the point cloud and its target, i.e., the corresponding 3D bounding box label.

To compare the data on a high level before training object detection algorithms, we calculate multiple statistical parameters of each dataset and compare them cross-dataset.
These statistical parameters contain the mean, minimum and maximum values of the point cloud ranges, the number of points per entire point cloud, and the number of points per target bounding box.

\subsection{Object Detection Networks and Configuration}
\label{sec:method_networks}

To quantify the dataset similarity and performance of the re-simulated data evaluated on the \textit{real} dataset, we use state-of-the-art 3D object detection algorithms.
Object detection on point clouds can be categorized into point-based and voxel-based approaches.
As these approaches might behave differently to a domain shift, we choose one algorithm for each category.
Note that our goal is not to compare the network's performance against each other but to compare the datasets by evaluating each network individually.
For the voxel-based approach, we choose PointPillars \cite{Lang2019PointPillars:Clouds}, which extracts features from vertical columns of the point cloud to predict 3D bounding boxes of the objects.
PointRCNN \cite{Shi2018PointRCNN:Cloud} is our choice for a point-based object detection algorithm, which uses PointNet++ \cite{Qi2017PointNet++:Space} as a backbone to extract local features at the point-level.

As our datasets contain only one object class that needs to be detected, we adapt the output layers of both algorithms to predict a single class only.
The anchor size is set to the ground truth dimensions of the race car, with $l = \SI{4.88}{\meter}$, $w = \SI{1.90}{\meter}$, and $h = \SI{1.18}{\meter}$ for length $l$, width $w$, and height $h$, respectively.
We empirically test different network parameters, such as the voxel size for PointPillars, the number of sampled points in the feature extractor of PointRCNN, or the number of filters in both networks.
For all other parameters, we use the default values for each network.
Although the LiDAR sensors capture reflections at over \SI{200}{\meter} distance, we limit the detection range of our networks to a horizontal range of \SI{100}{\meter} in the dimensions x and y.
We remove the intensity channel from both networks and only use $x$, $y$, and $z$ as input features.

Each combination of network and dataset is trained for $75$ epochs, after which no further decrease in validation loss is found.
In general, the network training is non-deterministic, whereas the non-determinism is more pronounced at \mbox{PointRCNN}.
Therefore, we train every network and dataset configuration with identical parameters five times.
We report the mean and standard deviation for the selected KPIs (see Sec.~\ref{sec:method_kpis}) in the results.

\subsection{KPIs for Domain Shift Quantification}
\label{sec:method_kpis}

For the quantitative evaluation of the object detection algorithms presented, we need a metric that assesses the performance of each trained network and dataset configuration.
This metric should be based on the network's final outputs, i.e., the predicted 3D bounding boxes.
We use average precision (AP), which is a standard metric for object detection.
This metric compares the predicted 3D bounding boxes with the ground truth 3D bounding boxes and classifies each predicted box into a true positive (TP) or false positive (FP) based on the 3D overlap with the ground truth boxes.
If a predicted bounding box reaches a certain threshold of intersection over union (IoU) with a ground truth box, it is classified as TP, otherwise as FP.
Based on the TP, FP, and missed ground truth boxes, i.e., false negatives (FN), the recall and precision of the network and dataset configuration can be calculated.
The final AP is the area under the precision-recall-curve; more precisely, we use the 40-point interpolated AP as described in \cite{Simonelli2019DisentanglingDetection}.
We report the AP for two different IoU thresholds, \SI{50}{\percent} and \SI{70}{\percent} overlap, denoted as 3D AP (0.5) and 3D AP (0.7), respectively.

\section{Results}
\label{sec:results}

\begin{table}[!t]
\centering
\caption{Statistical comparison of general aspects of the real and simulated data. All values are calculated based on the full training set of each dataset, using the entire point cloud range up to 100 m.}
\label{tab:stat_comp}
\resizebox{\columnwidth}{!}{%
\begin{tabular}{lllrr}
\hline
\multicolumn{3}{c}{Attribute} &
  \multicolumn{1}{c}{Real-world data} &
  \multicolumn{1}{c}{Simulated data} \\ \hline
\multirow{9}{*}{\begin{tabular}[c]{@{}l@{}}Point cloud range\\ in meters\end{tabular}} &
  \multirow{3}{*}{mean} &
  $x$ &
  $2.2$ &
  $2.1$ \\
 &                      & $y$ & $-1.3$   & $-0.5$   \\
 &                      & $z$ & $1.2$    & $1.9$    \\ \cline{2-5} 
 & \multirow{3}{*}{min} & $x$ & $-100.0$ & $-100.0$ \\
 &                      & $y$ & $-100.0$ & $-87.0$  \\
 &                      & $z$ & $-12.0$  & $-1.5$  \\ \cline{2-5} 
 & \multirow{3}{*}{max} & $x$ & $100.0$  & $100.0$  \\
 &                      & $y$ & $100.0$  & $100.0$  \\
 &                      & $z$ & $25.1$   & $23.3$   \\ \hline
\multirow{3}{*}{\begin{tabular}[c]{@{}l@{}}Number of points\\ per point cloud\end{tabular}} &
  mean &
   &
  $73{,}123$ &
  $78{,}776$ \\
 & min                  &   & $52{,}762$  & $52{,}695$  \\
 & max                  &   & $79{,}690$  & $81{,}538$  \\ \hline
\multirow{3}{*}{\begin{tabular}[c]{@{}l@{}}Number of points\\ per target box\end{tabular}} &
  mean &
   &
  $219$ &
  $251$ \\
 & min                  &   & $0$      & $5$      \\
 & max                  &   & $4{,}959$   & $6{,}465$   \\ \hline
\end{tabular}%
}
\end{table}

Following the method presented in Sec.~\ref{sec:method}, this section starts by presenting the results of the statistical dataset comparison and the quantification of the domain shift using object detection algorithms in Sec.~\ref{sec:results_stat_data_comp} and Sec.~\ref{sec:results_obj_det}, respectively.
This is followed by a detailed point cloud target-level analysis in Sec.~\ref{sec:results_target_analysis} and the following additional study in Sec.~\ref{sec:4d}.

\subsection{Statistical Dataset Comparison}
\label{sec:results_stat_data_comp}

\begin{figure}[t!]
	\centering
	\subfloat{{\includegraphics[width=0.9\linewidth]{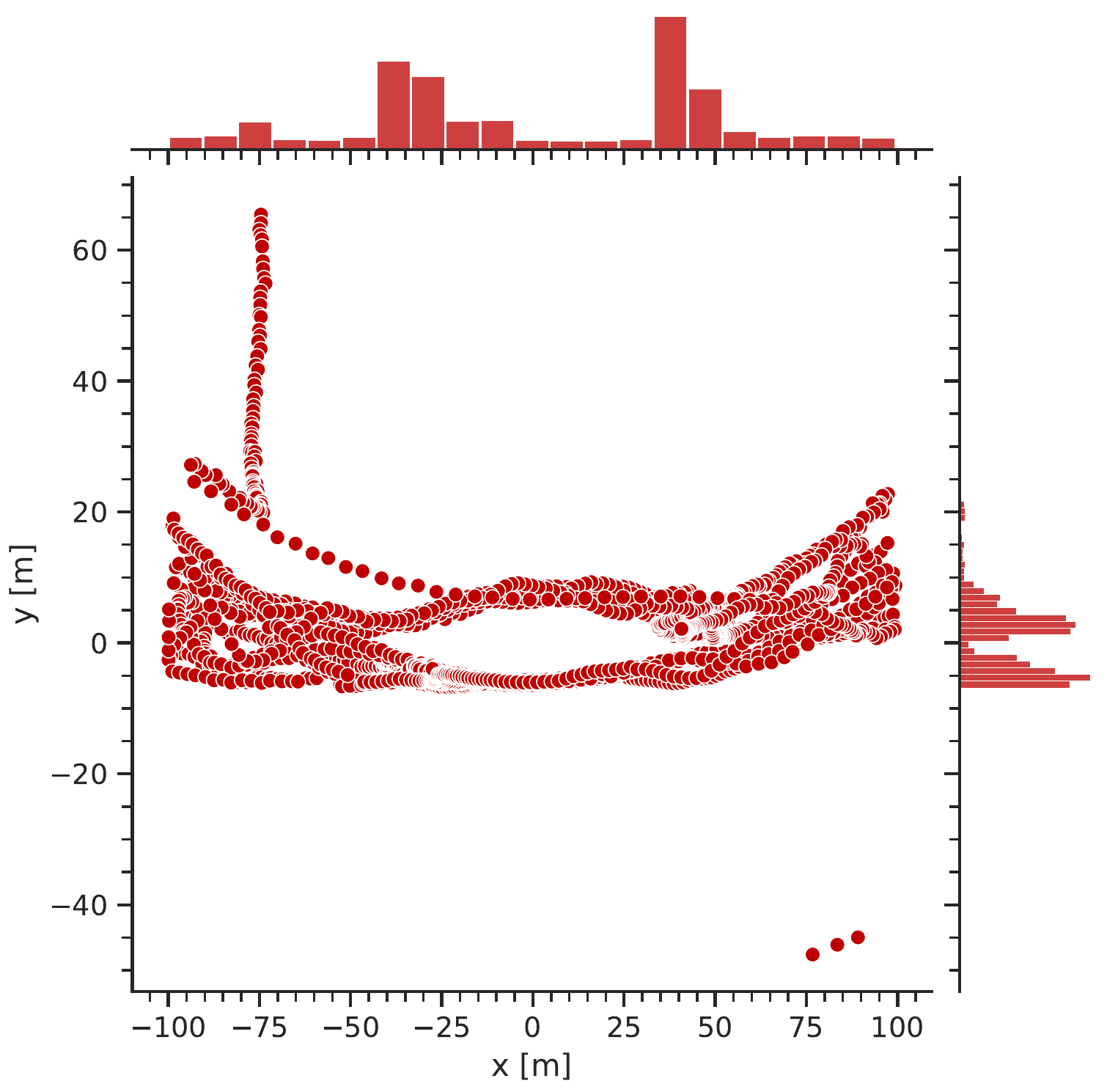} }}
	\caption{Location of the target vehicle relative to the ego vehicle for all samples in the \textit{real} dataset. The histograms on the axes show the distribution of the depicted locations.}\label{fig:rel_locations}%
\end{figure}

\begin{figure*}[t!]
	\centering
	\fontsize{7pt}{6pt}
	\captionsetup{format=hang, justification=centering, indention=0.55cm}
	\subfloat[{Full Range $\left[\SI{0.0}{\meter}, \SI{100.0}{\meter}\right]$}\label{fig:PointRCNN_AP_0_7_range_based_full}]{\selectfont\def\svgwidth{0.24\linewidth}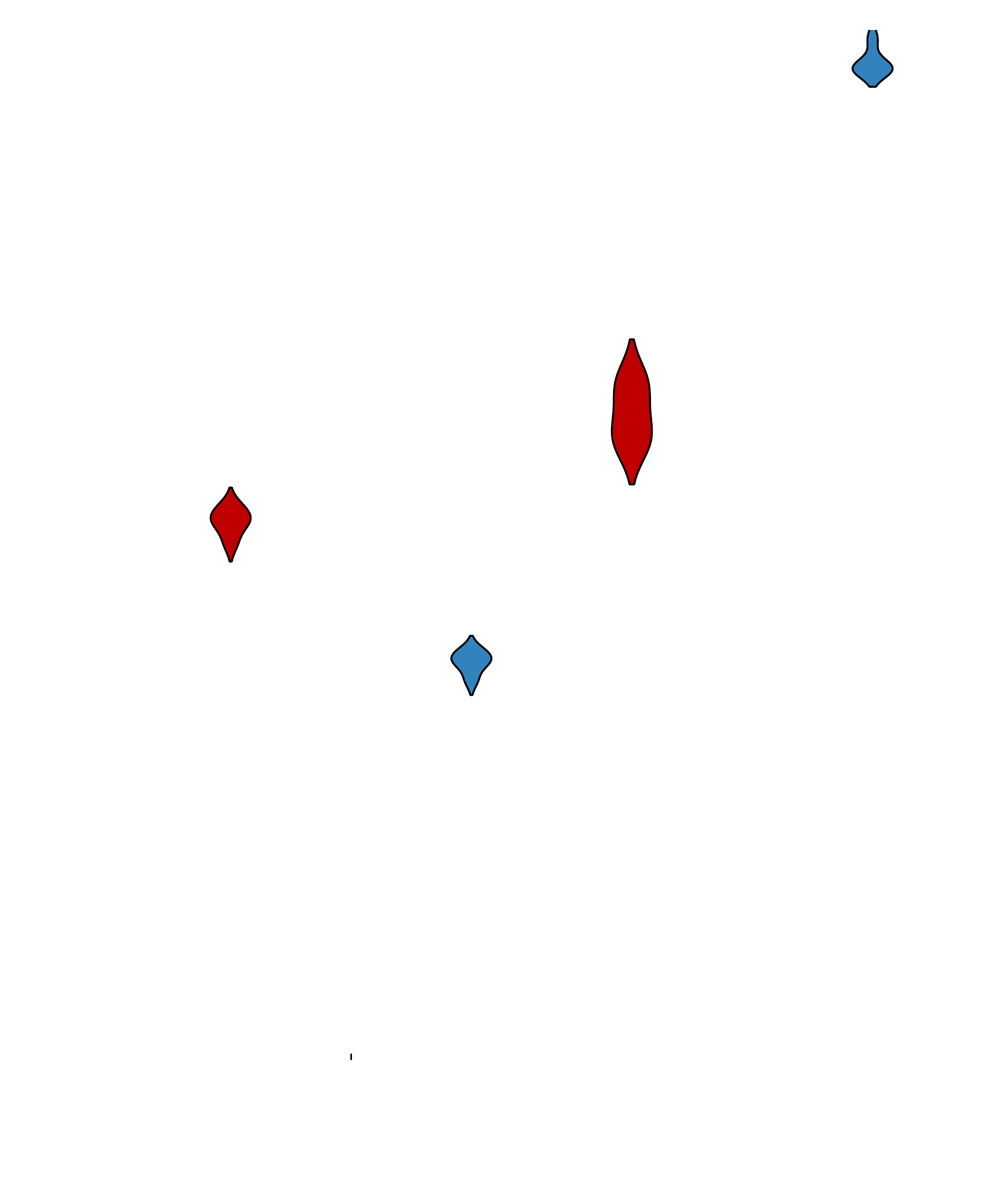}
	\hfill
	\captionsetup{format=hang, justification=centering, indention=0.1cm}
	\subfloat[{Close-Range $\left[\SI{0.0}{\meter}, \SI{33.3}{\meter}\right[$}\label{fig:PointRCNN_AP_0_7_range_based_close}]{\selectfont\def\svgwidth{0.24\linewidth}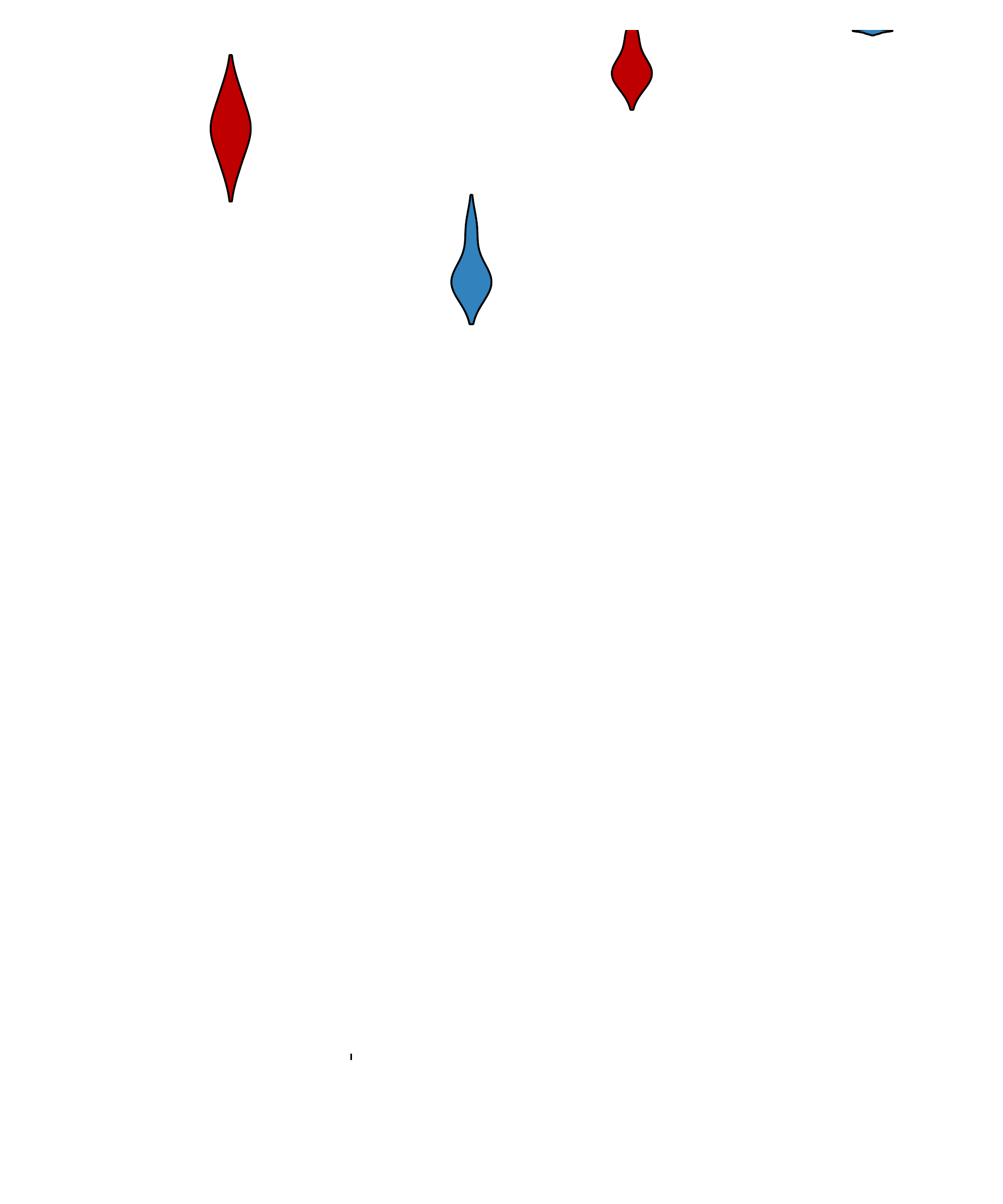}
	\hfill
	\captionsetup{format=hang, justification=centering, indention=0.55cm}	
	\subfloat[{Mid-Range $\left[\SI{33.3}{\meter}, \SI{66.6}{\meter}\right[$}\label{fig:PointRCNN_AP_0_7_range_based_mid}]{\selectfont\def\svgwidth{0.24\linewidth}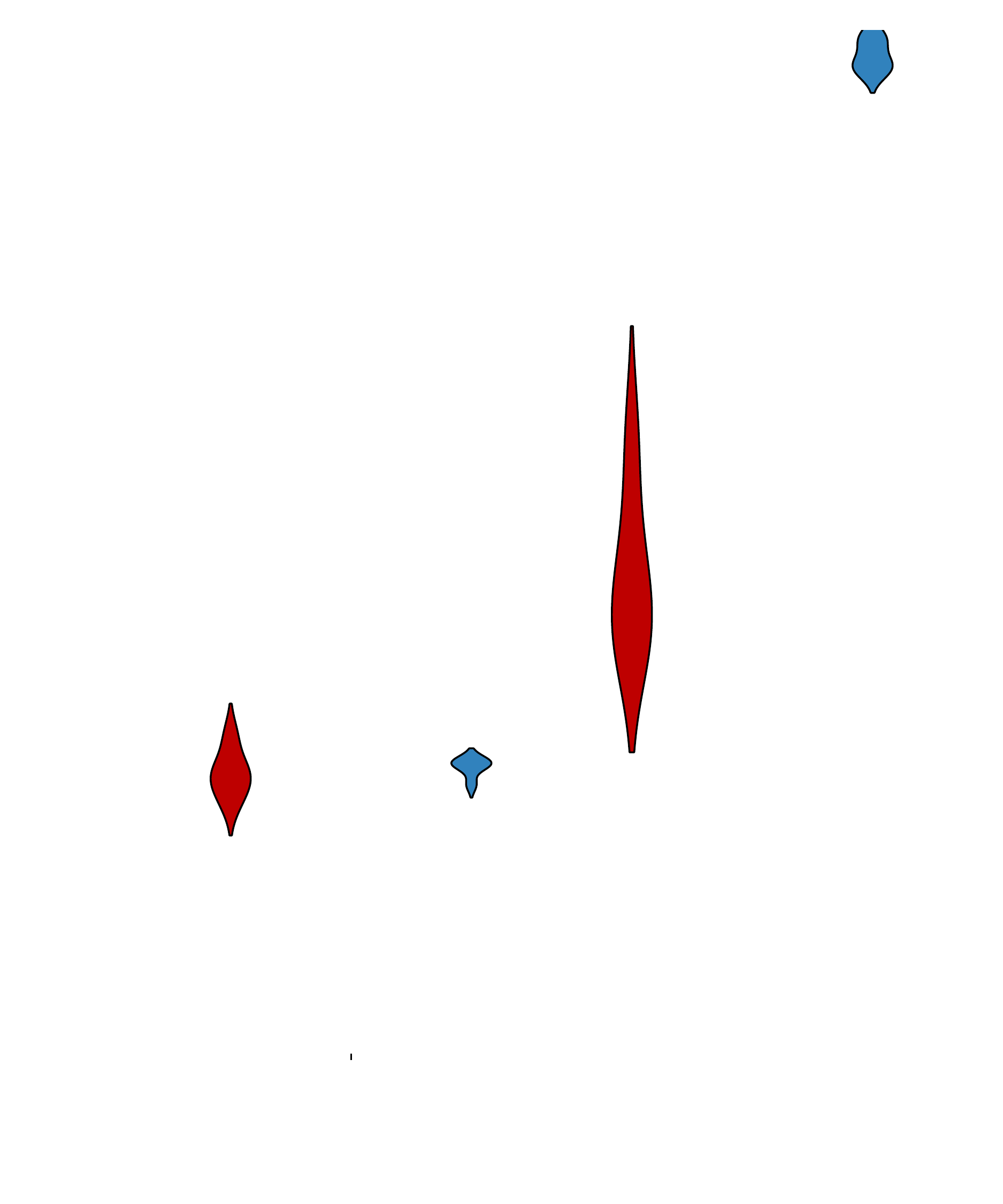}
	\hfill
	\captionsetup{format=hang, justification=centering, indention=0.5cm}	
	\subfloat[{Long-Range $\left[\SI{66.6}{\meter}, \SI{100.0}{\meter}\right]$}\label{fig:PointRCNN_AP_0_7_range_based_long}]{\selectfont\def\svgwidth{0.24\linewidth}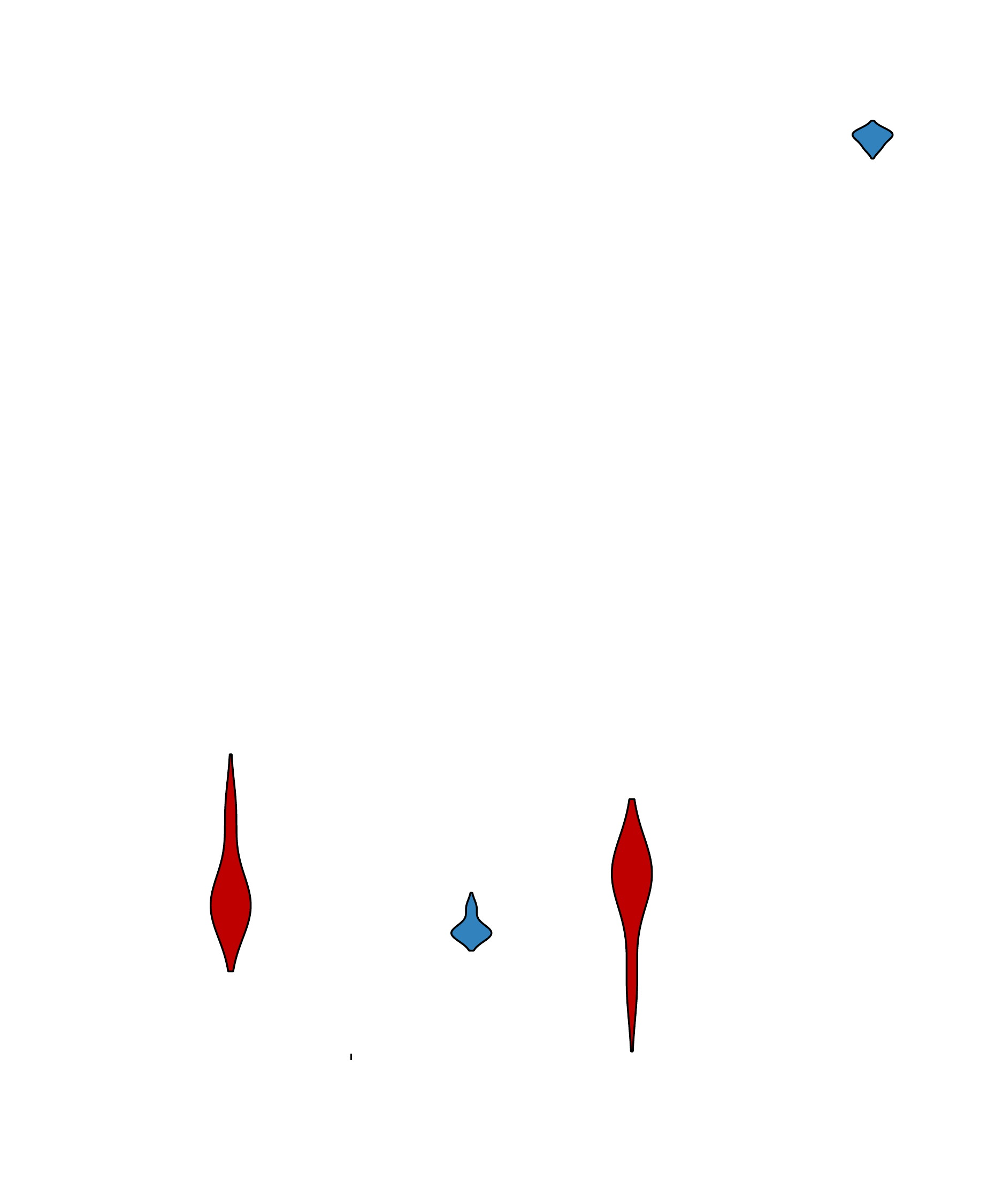}
	\caption{Average precision with IoU \SI{70}{\percent} for PointRCNN trained and tested with \textit{real} or \textit{sim} data. The horizontal lines mark the mean AP, and the five points mark the individual five training runs per train-test pairing. Note that at (b), the mean AP of sim-to-sim is close to \SI{100}{\percent} and, therefore, barely visible.}\label{fig:PointRCNN_AP_0_7_range_based}
\end{figure*}

Table~\ref{tab:stat_comp} provides an overview of the statistical metrics selected for the comparison of the datasets.
By comparing the point cloud range of \textit{sim} and \textit{real}, it can be seen that in the simulation, certain locations are not hit, which are hit in reality by the LiDAR.
An example is the y coordinate of the minimum ($-\SI{100}{\meter}$ vs. $-\SI{87}{\meter}$). 
A model trained on the \textit{sim} data will never encounter points in certain locations that are included in the \textit{real} dataset.
Note that none of these missing locations are on the race track, which means that the trained model will not see a single instance where such outside points contain a bounding box of a target.

Generally, the LiDAR used is capable of hitting and measuring objects further away than \SI{100}{\meter}, but from our observations, these long-distance measurements mostly resemble noise, making the additional computing effort to treat them not worth it. 
In addition, the number of measured points drops drastically in a higher range. 

One of the most noticeable differences is the number of points collected from the target. 
It is quite intuitive that with increasing distance of the target to the ego vehicle, the number of points drops, i.e., the density of the target point cloud is reduced.
This effect is less dominant within the \textit{sim} dataset, where by default no measurements are lost due to the increased distance. 
We can roughly estimate the loss ratio of the laser beams in the \textit{real} dataset by comparing the mean number of points per target box: $\frac{219}{251}~\approx~\SI{87}{\percent}$. 
There is even one instance where the target in the \textit{real} dataset is not hit entirely, which provides an interesting edge case. 

For both the \textit{real} and the \textit{sim} datasets, we only consider the heading angle (yaw) and neglect the roll and pitch angles, which is common in the domain of autonomous driving.
The heading angle of the vehicles in the simulation is taken from the \textit{real} data, and therefore they are identical for every sample pairing.
Overall, the vast majority of the samples in our dataset show target boxes with only a small relative heading angle.
Thus, the target is seen from mostly similar positions, but these small differences are enough for the LiDAR to capture points throughout the target vehicle.

From the birds-eye view of two corresponding representative images taken from the \textit{real} and the \textit{sim} datasets (Fig.~\ref{fig:full_pcd_bev}), it can be seen how similar the point clouds generally are.
A noticeable difference is a trapezoidal-shaped area (in the \textit{real} sample) behind the vehicle in the driving direction (the driving direction is to the right). 
This area in the scan is due to the placement of the LiDAR on top of the vehicle. 
This leads to a blind spot caused by the rear wing shielding part of the track from being reached by the laser beams, which can be seen in Fig.~\ref{fig:car_real}. 
This blind area is slightly larger than \SI{10}{\meter} long and \SI{5}{\meter} wide. 
A couple of similar artifacts resulting from the ego vehicle can be seen around the point cloud origin. 
None of these blind spots are problematic in this dataset since cases where the target is mostly hidden in this space do not exist; compare to Fig.~\ref{fig:rel_locations}, showing the target vehicle locations of the entire dataset.

Just as important as keeping an eye on the point clouds is to analyze the location distribution of the targets. 
This is relevant because a model trained with most targets in close proximity might struggle to generalize to targets farther away since the target point clouds follow a different and more inaccurate measured shape.
Fig.~\ref{fig:rel_locations} shows the location in the $x$-$y$-plane of the target vehicles for each sample in the \textit{real} dataset. 
Furthermore, we show the distribution of the target locations for the $x$- and $y$-axis.
Two main locations of the target vehicle are predominant: to the left front ($+\SI{35}{\meter}$, $+\SI{5}{\meter}$) and to the right back ($-\SI{30}{\meter}$, $-\SI{5}{\meter}$).

Although minor differences exist, we argue that the statistical dataset comparison shows an overall high similarity between the \textit{real} dataset and the derived \textit{sim} dataset.
Furthermore, the similarity of our scenario-identical datasets is higher than any publicly available dataset pair, which makes them suitable for the following detailed sim-to-real domain shift analysis. 

\vspace{-5pt}

\subsection{Object Detection Evaluation}
\label{sec:results_obj_det}
This section presents the quantitative results of object detection algorithms trained with \textit{real} or \textit{sim} datasets.
To account for the non-deterministic training, each network and dataset pair is trained five times with identical configurations, as stated in Sec.~\ref{sec:method_networks}.
We report the mean and standard deviation of the calculated AP of the five training runs for each network and dataset pair.
A table presenting all results for PointRCNN and PointPillars can be found in Table~\ref{tab:all_results} in Appendix \ref{appendix:results}.
In the following, our notation is as follows: An experiment with a network trained on \textit{real} data and evaluated on \textit{sim} data is noted as ``real-to-sim'', i.e., the first word describes the training dataset, and the last word describes the test dataset.

Fig.~\ref{fig:PointRCNN_AP_0_7_range_based_full} shows the 3D AP (0.7) for PointRCNN for all four possible pairings of \textit{real} and \textit{sim} data for training and testing.
In this experiment, training and testing are conducted on the full range, including targets up to a range of \SI{100}{\meter}.
The performance comparison of sim-to-real (\SI{38.23}{\percent}) and real-to-real (\SI{51.96}{\percent}) indicates the existence of a distinct sim-to-real domain shift.
Although the network was trained with exactly the same scenarios and target distributions, performance drops by almost \SI{14}{\percent} just by training with the \textit{sim} data.

Lower performance of the opposite domain can also be observed when the networks are tested on \textit{sim} data.
Comparing sim-to-sim (\SI{96.82}{\percent}) with real-to-sim (\SI{62.53}{\percent}), the real-to-sim domain shift of around \SI{34}{\percent} is even more pronounced than the sim-to-real domain shift (\SI{14}{\percent}).
This can be explained by the overall higher performance of sim-to-sim (\SI{96.82}{\percent}) compared to real-to-real (\SI{51.96}{\percent}).
The behavior of a higher real-to-sim domain shift compared to the sim-to-real domain shift is typical for simulated data and agrees with the results of \cite{Dworak2019PerformanceSimulator} and \cite{Nowruzi2019HowData}.

The results for PointRCNN are consistent with the results of PointPillars.
In general, PointRCNN achieves higher AP by a large margin in most train-test pairings, except that PointPillars outperforms PointRCNN in the sim-to-sim pairing.
However, as stated before, this pairing leads to an almost perfect AP, independent of the network choice.

As expected, lowering the IoU threshold for the AP calculation from \SI{70}{\percent} to \SI{50}{\percent} leads to higher APs overall, without exception.
The absolute AP difference between IoU \SI{70}{\percent} and \SI{50}{\percent} is lowest when the AP is close to \SI{100}{\percent}, i.e., for sim-to-sim pairings.

\begin{figure}[t!]
	\centering
	\subfloat{\includegraphics[width=0.9\columnwidth]{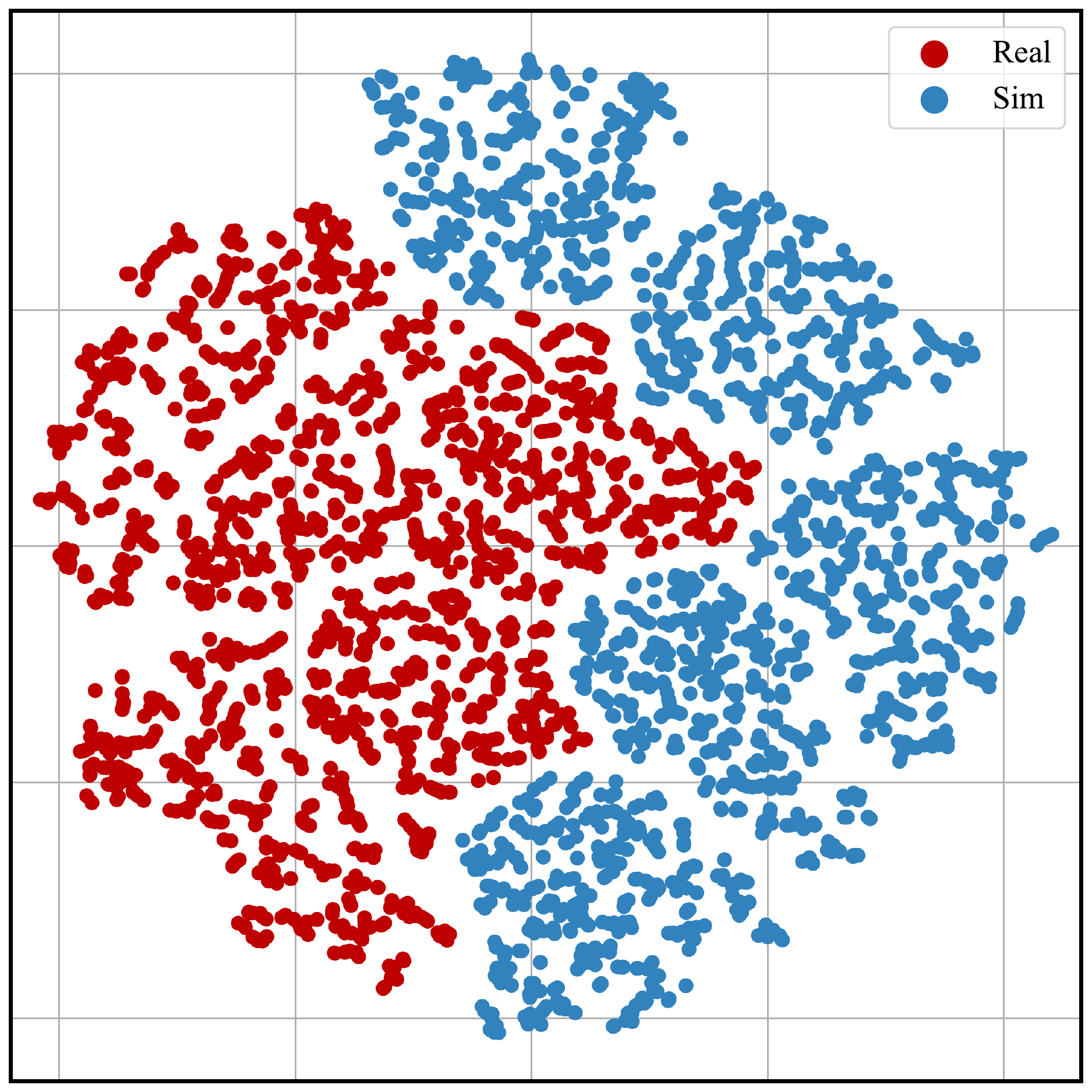}}
	\caption{T-SNE of the latent feature space of PointRCNN trained on \textit{real} or \textit{sim} data for five training runs each. Each point visualizes a feature vector generated by network inference with a single point cloud.}\label{fig:tsne_real_sim}%
\end{figure}

\begin{figure*}[t!]
	\centering
        \captionsetup{format=hang}
	\subfloat[{Real, Close-Range $\left[\SI{0.0}{\metre}, \SI{33.3}{\metre}\right[$}\label{fig:car_pcd_real_0}]{\includegraphics[width=0.33\linewidth]{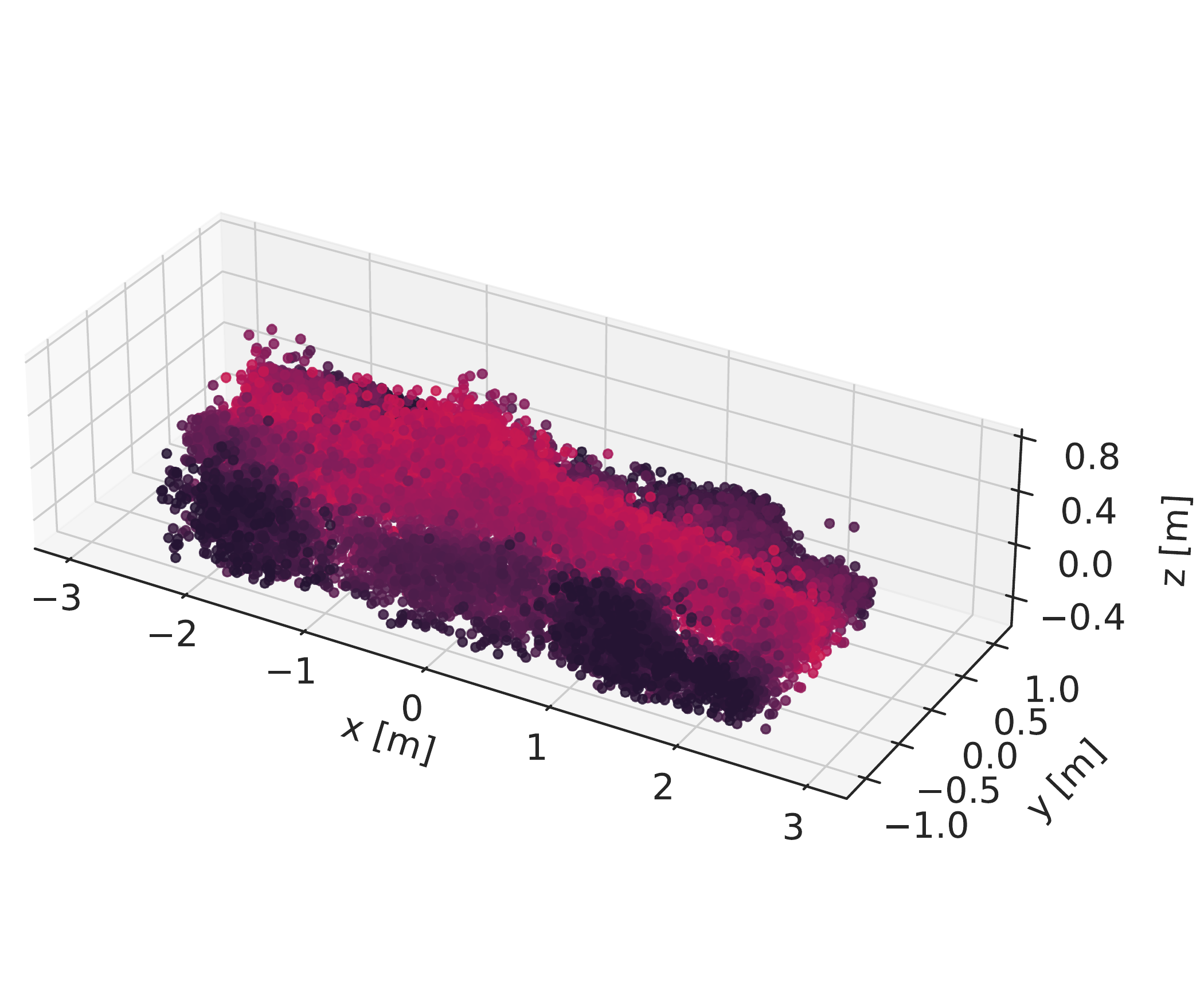}}
	\hfill
	\subfloat[{Real, Mid-Range $\left[\SI{33.3}{\metre}, \SI{66.6}{\metre}\right[$}\label{fig:car_pcd_real_33}]{\includegraphics[width=0.33\linewidth]{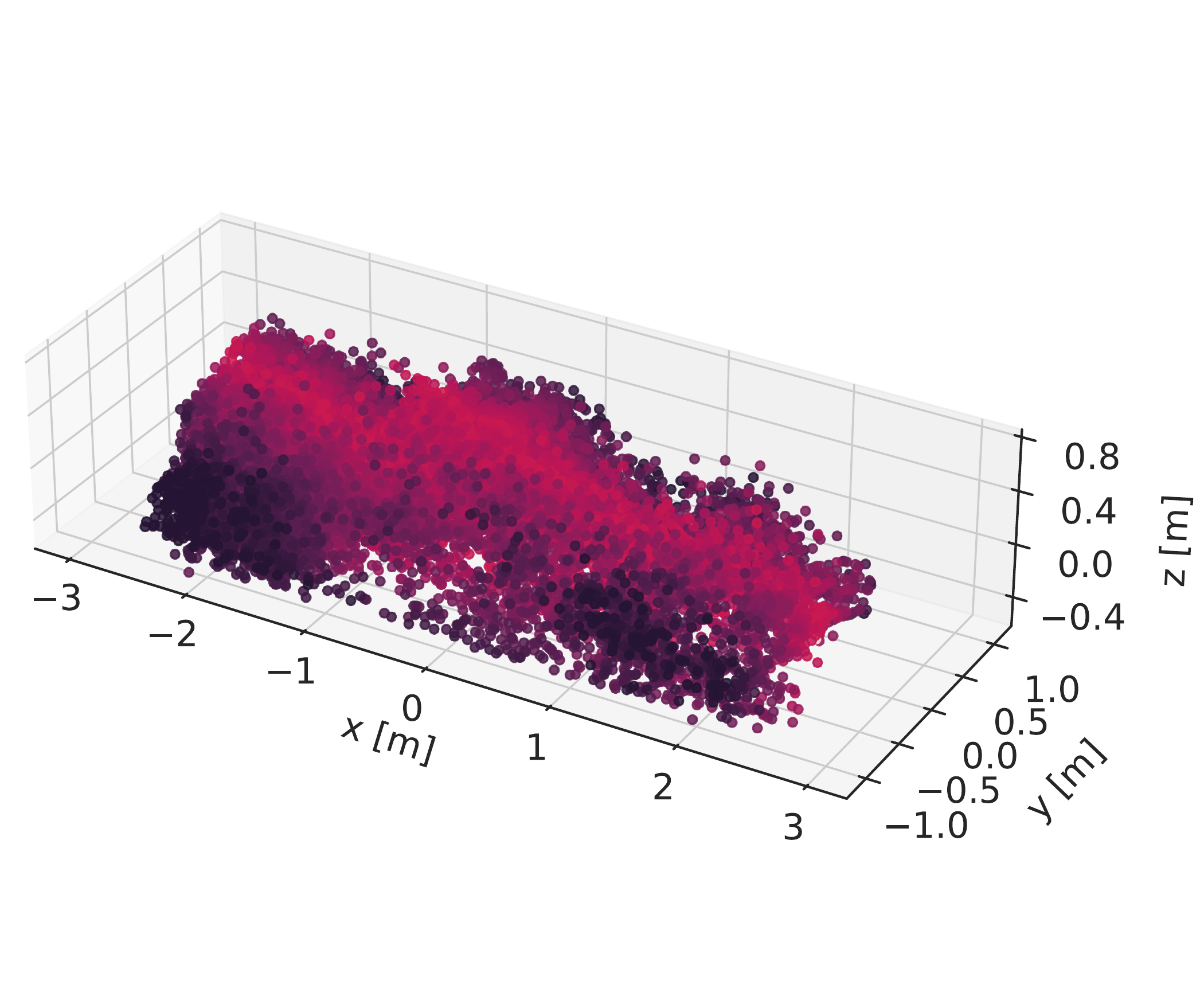}}
	\hfill
	\subfloat[{Real, Long-Range $\left[\SI{66.6}{\metre}, \SI{100.0}{\metre}\right]$}\label{fig:car_pcd_real_66}]{\includegraphics[width=0.33\linewidth]{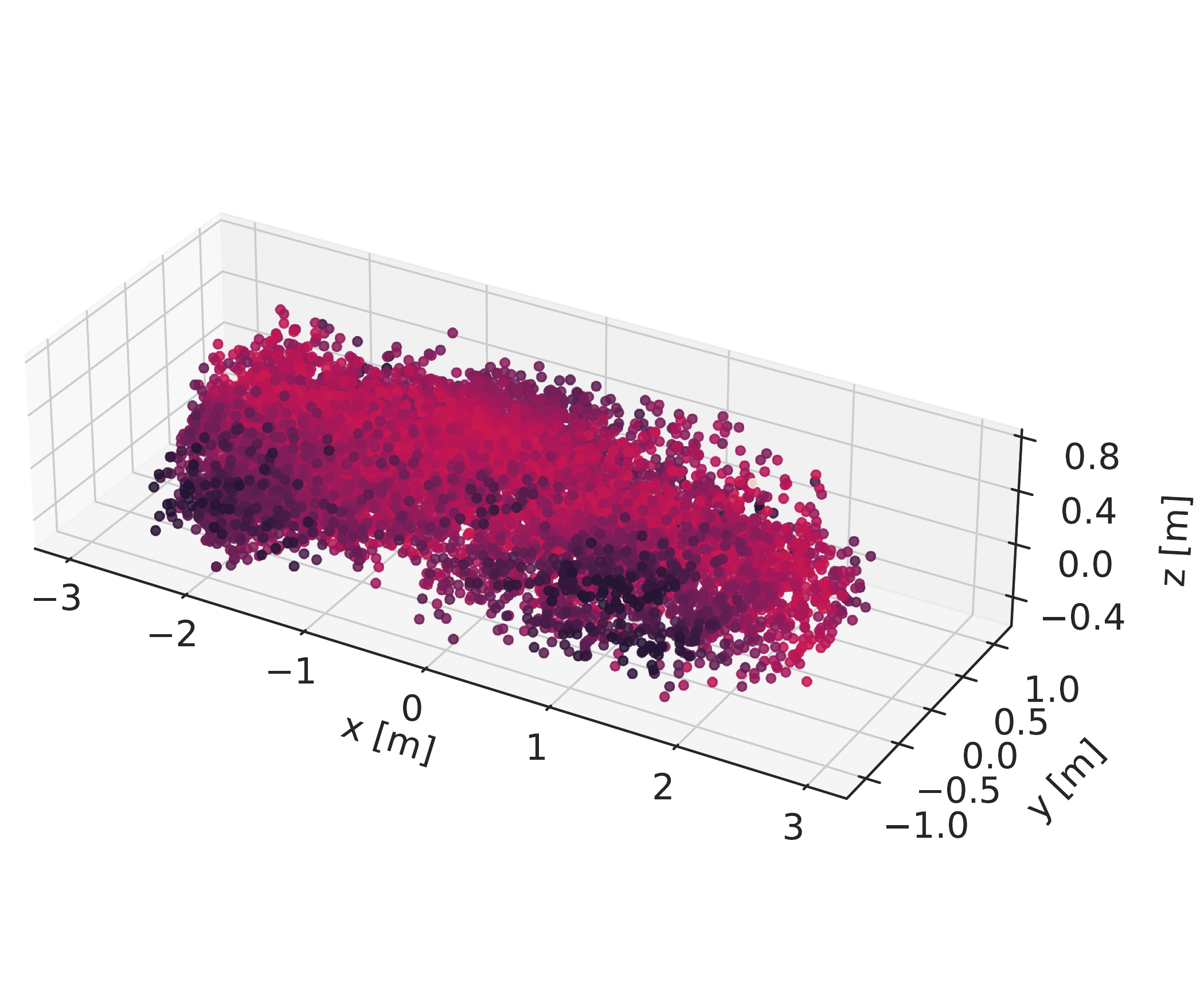}}

	\subfloat[{Sim, Close-Range $\left[\SI{0.0}{\metre}, \SI{33.3}{\metre}\right[$}\label{fig:car_pcd_sim_0}]{\includegraphics[width=0.33\linewidth]{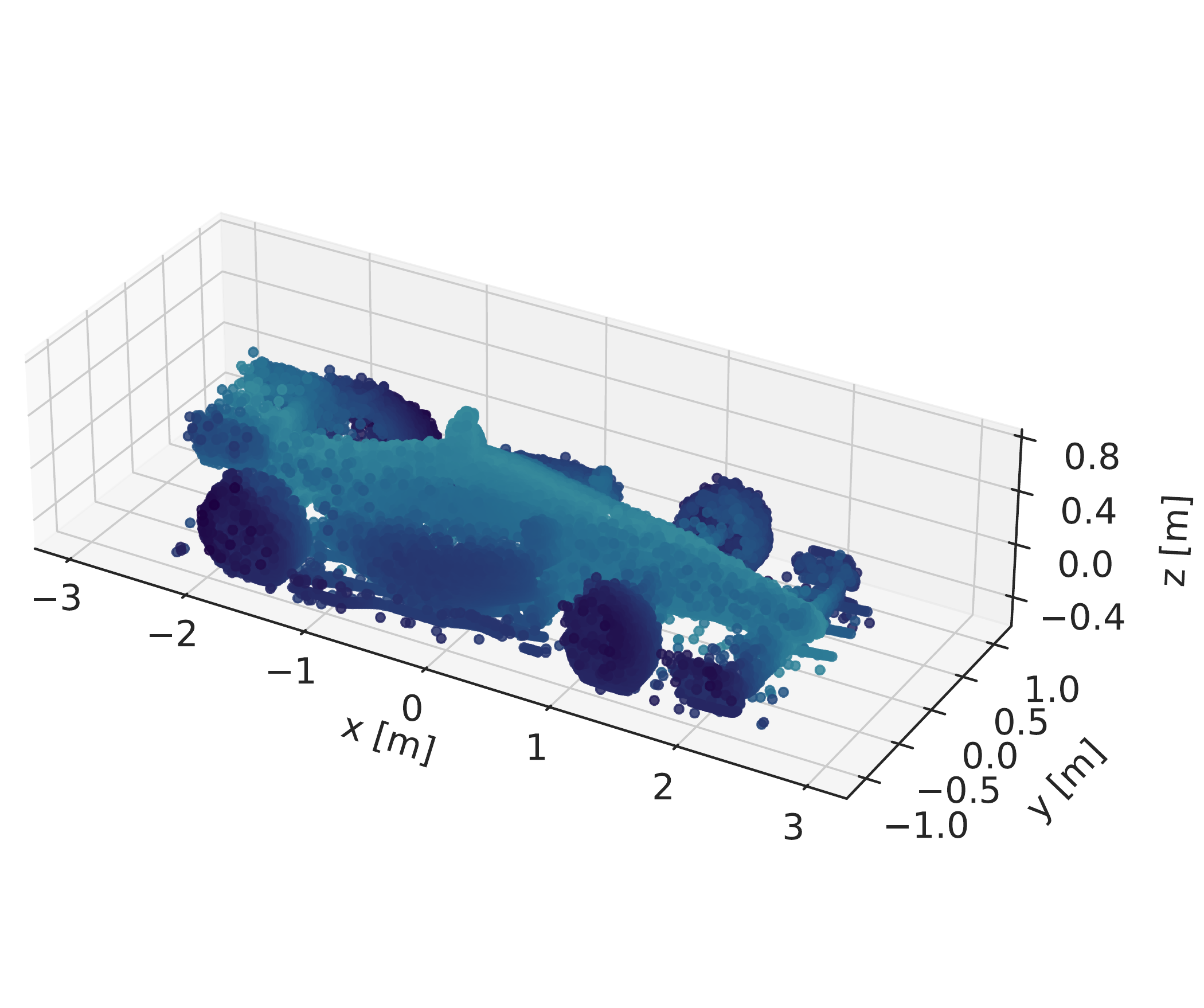}}
	\hfill
	\subfloat[{Sim, Mid-Range $\left[\SI{33.3}{\metre}, \SI{66.6}{\metre}\right[$}\label{fig:car_pcd_sim_33}]{\includegraphics[width=0.33\linewidth]{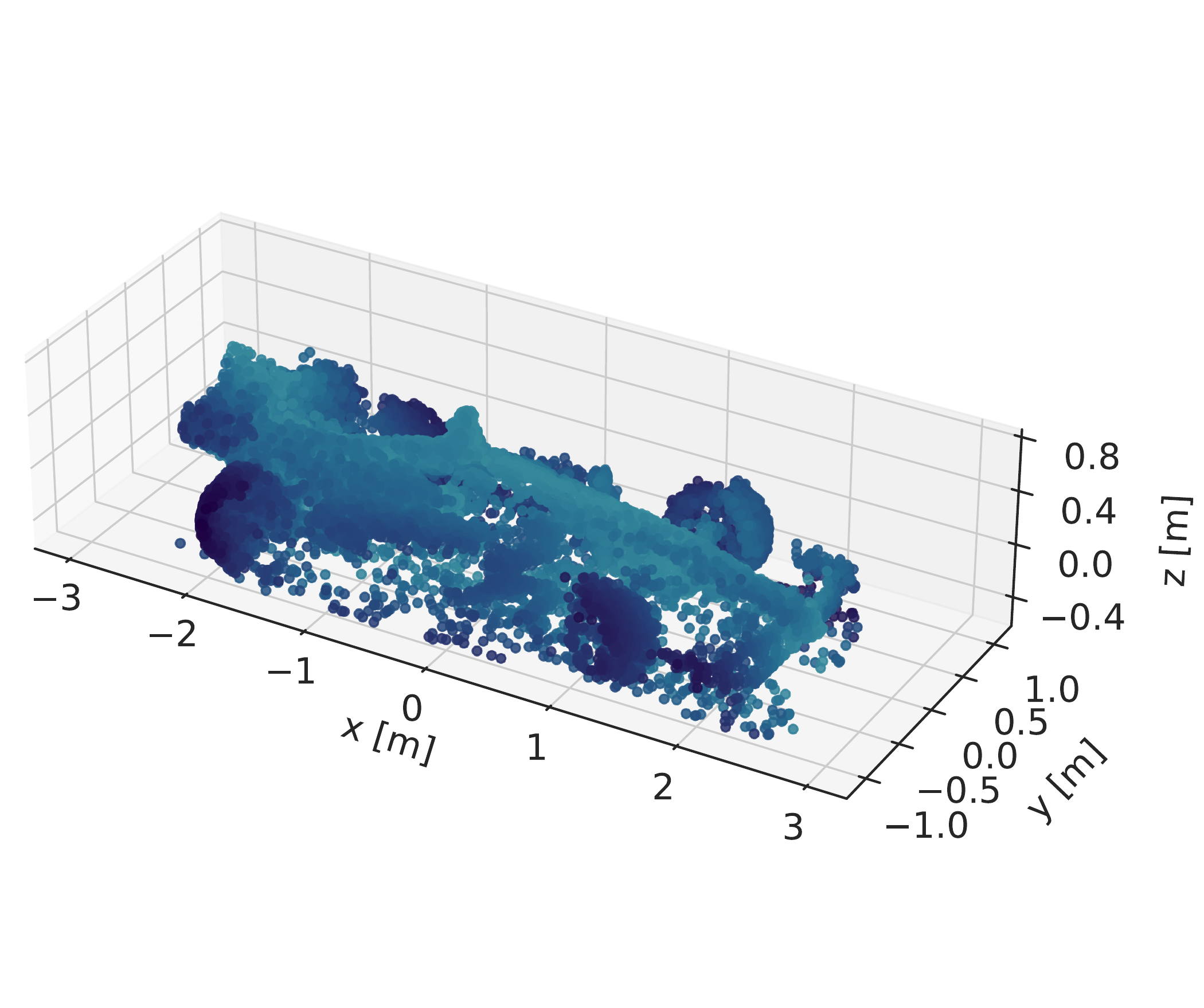}}
	\hfill
	\subfloat[{Sim, Long-Range $\left[\SI{66.6}{\metre}, \SI{100.0}{\metre}\right]$}\label{fig:car_pcd_sim_66}]{\includegraphics[width=0.33\linewidth]{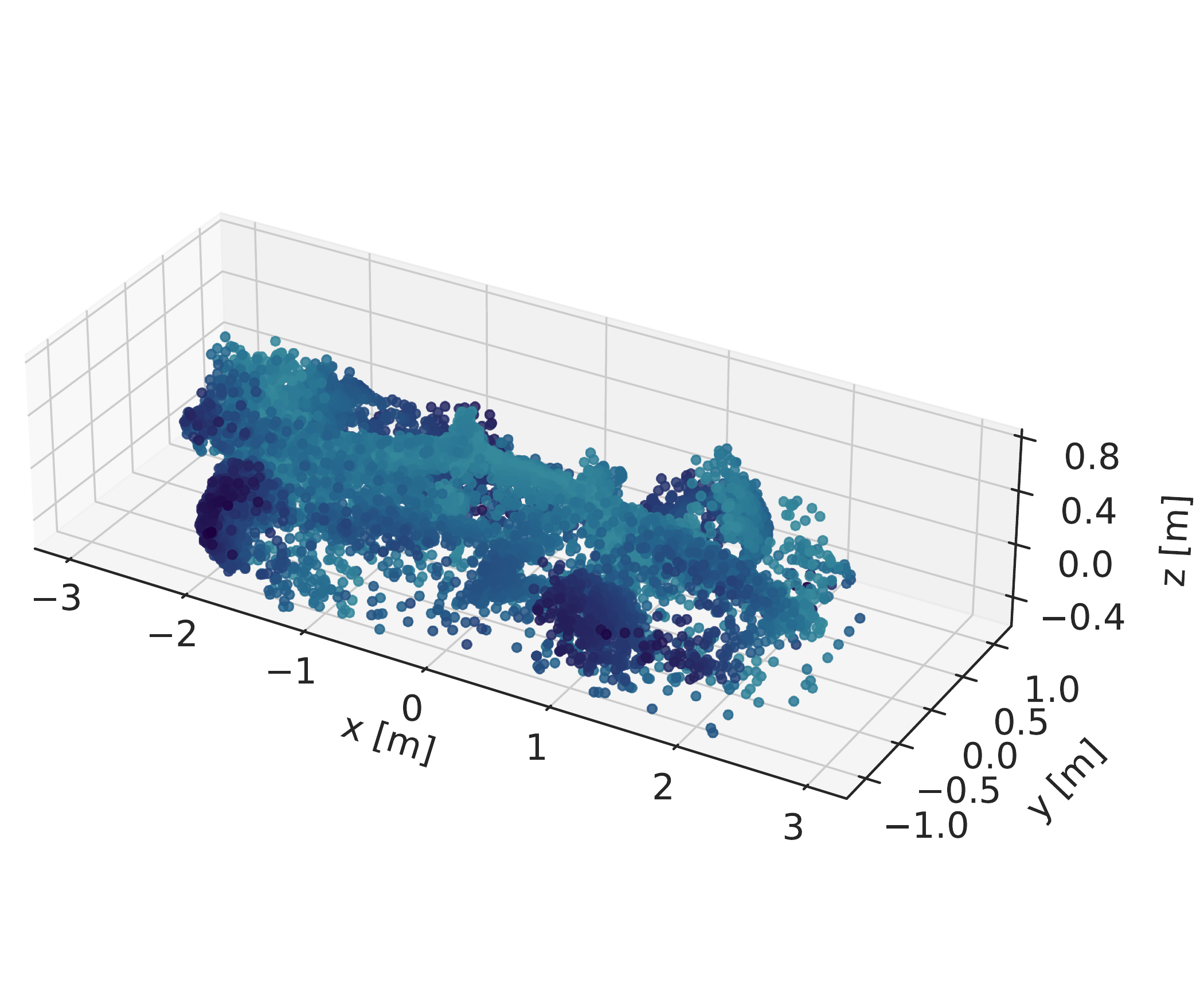}}

	\caption{Aggregated normalized point clouds from the real dataset (first row) and sim dataset (second row). The columns represent the distance from the target before normalization: close-range \SI{0.0}{\meter} to \SI{33.3}{\meter} (a)(d), mid-range \SI{33.3}{\meter} to \SI{66.6}{\meter} (b)(e), and long-range \SI{66.6}{\meter} to \SI{100}{\meter} (c)(f). We randomly select $20{,}000$ points from all measured target points for close- and mid-range and $9{,}732$ for long-range (all available points of targets in the real dataset at long-range) for visualization purposes.}\label{fig:car_pcd_overview}
\end{figure*}

We further analyze the domain shift by evaluating the networks in close-range $r_1$, mid-range $r_2$, and long-range $r_3$ as defined in Sec.~\ref{sec:method_stat_data_comp}.
The results showing 3D AP (0.7) for PointRCNN for these three ranges are depicted in Fig.~\ref{fig:PointRCNN_AP_0_7_range_based_close}-\ref{fig:PointRCNN_AP_0_7_range_based_long}.
As expected, overall performance decreases with an increasing range, an observation that is valid for each of the four train-test pairings.
The loss of performance is most pronounced when the close-range is compared with the mid-range.
It is interesting to note that sim-to-sim performance only decreases slightly from close-range (\SI{99.88}{\percent}) to long-range (\SI{89.47}{\percent}), whereas the real-to-real performance drops from \SI{90.41}{\percent} to \SI{16.31}{\percent}.

Another observation concerns the sim-to-real domain shift:
In the mid-range, the performance of sim-to-real and real-to-real is almost identical; hence, no sim-to-real domain shift can be observed.
However, in close-range and long-range, there is a sim-to-real domain shift, leading to the general sim-to-real domain shift in the combined full range \ref{fig:PointRCNN_AP_0_7_range_based_full}.

Fig.~\ref{fig:tsne_real_sim} shows the t-Distributed Stochastic Neighbor Embedding (t-SNE) of the high-dimensional latent feature space of PointRCNN trained on \textit{real} or \textit{sim} data.
T-SNE is a statistical method for the reduction of dimensionality of high-dimensional data and is suitable for the visualization of these data in two-dimensional plots \cite{VanDerMaaten2008VisualizingT-SNE}.
Neighboring points in the low-dimensional representation are usually similar in the high-dimensional input space.
Each point represents one feature vector generated by the inference of PointRCNN with one point cloud of the \textit{real} test set.
In total, the plot shows $2{,}500$ points for each dataset used for training, that is, the \textit{real} or \textit{sim} dataset.
Within each dataset, five clusters can be identified that originate from the five training runs of each configuration.
The plot indicates that the feature vectors are distinct, and hence, the network learned the different distributions of the \textit{real} and \textit{sim} data.

\subsection{Point Cloud Target-Level Analysis}
\label{sec:results_target_analysis}

On the basis of the quantitative results, we further qualitatively analyze the domain shift based on the point clouds of the targets.
In the datasets used, the target class is a single non-deformable object.
This enables visualization of the target shape the network will see during the training by aggregating the point clouds at the target-level.
If the aggregated shape is clearly distinguishable from the environment, the object detection network should detect it with high accuracy.

Fig.~\ref{fig:car_pcd_overview} shows that with increasing distance between the LiDAR and the target, the aggregated point clouds are dominated by noise. 
It is relevant for object detection to have enough points that follow the shape of the target. 
Especially in the \textit{real} dataset, long-range points become overly noisy, to the point that the target shape is barely recognizable anymore; compare Fig.~\ref{fig:car_pcd_real_66}.
A hard-to-recognize shape is per se not problematic if the relative difference to the environment is big enough to still be able to infer its existence. 
This becomes especially challenging for object detection algorithms without an attention mechanism that might hint at the most likely subsequent location (e.g., \cite{Wang2021AttentionCorrelation} or \cite{Liu2021Multi-objectMinimization}). 
Note also that distance-dependent noise due to outlier points is present in the simulated data, although it is less dominant.
However, the difference in noise is distributed as part of the domain shift (real-to-sim and sim-to-real), as this implies an additional generalization task for the model.

The reflections at long-range are mostly of the back of the vehicle due to the imbalance of target vehicle locations (compare Fig.~\ref{fig:rel_locations}).
This does not affect the domain shift, as it occurs equally in the \textit{real} and \textit{sim} datasets.
Still, this explains the back-heavy point clouds in Fig.~\ref{fig:car_pcd_real_66} and Fig.~\ref{fig:car_pcd_sim_66}.

\begin{figure}[t!]
	\subfloat[\centering Side view]{{\includegraphics[width=1\linewidth]{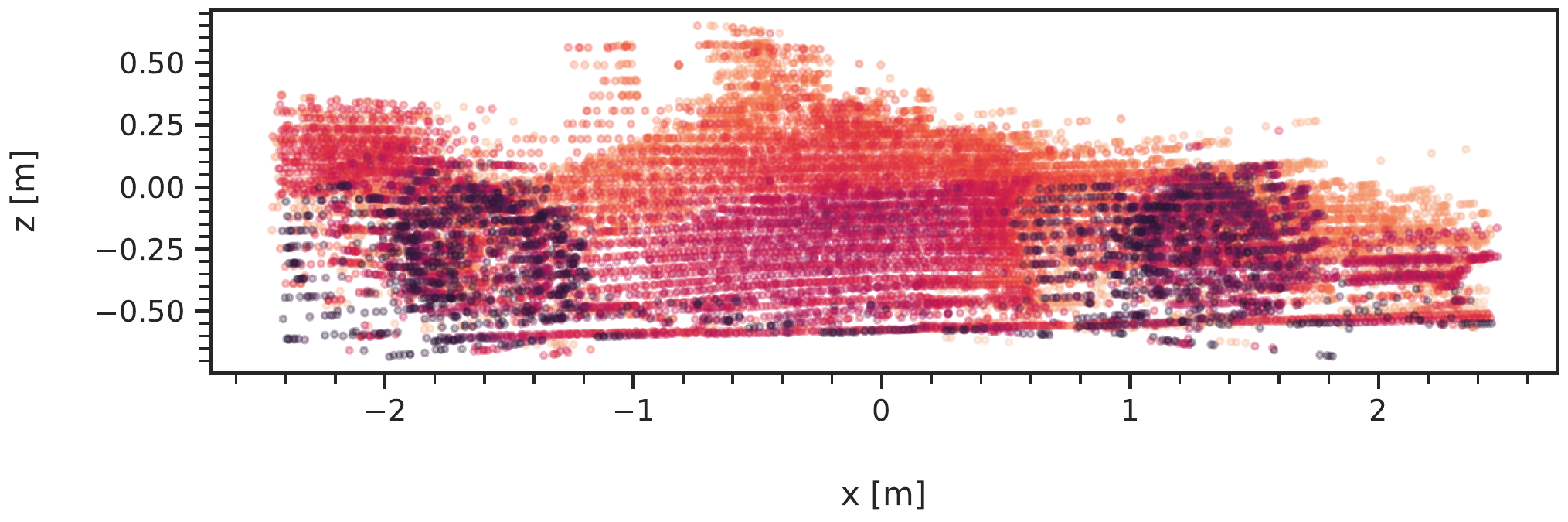}\label{fig:angle_view_side}}}

	\subfloat[\centering Top view]{{\includegraphics[width=1\linewidth]{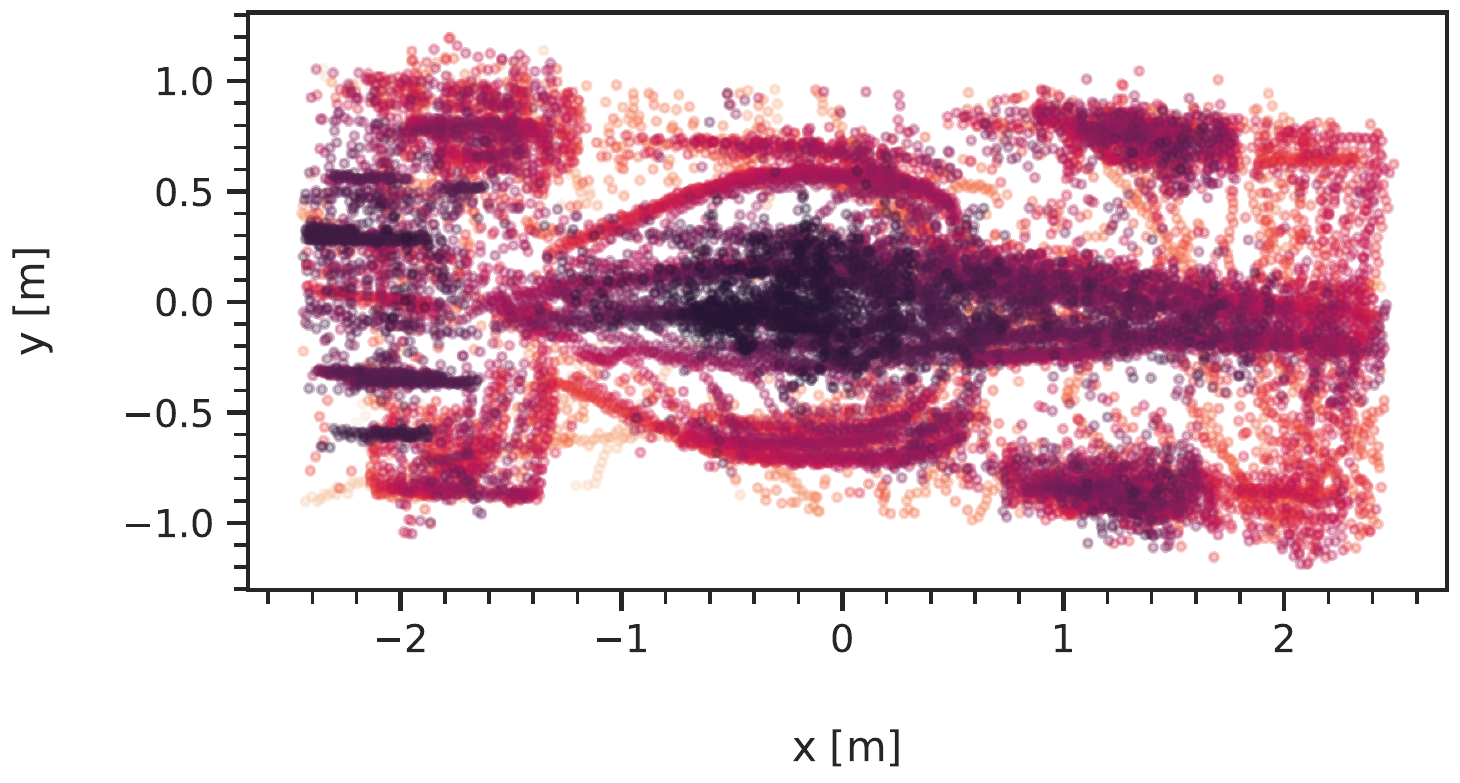}\label{fig:angle_view_top}}}

	\subfloat[\centering Front view]{{\includegraphics[width=0.48\linewidth]{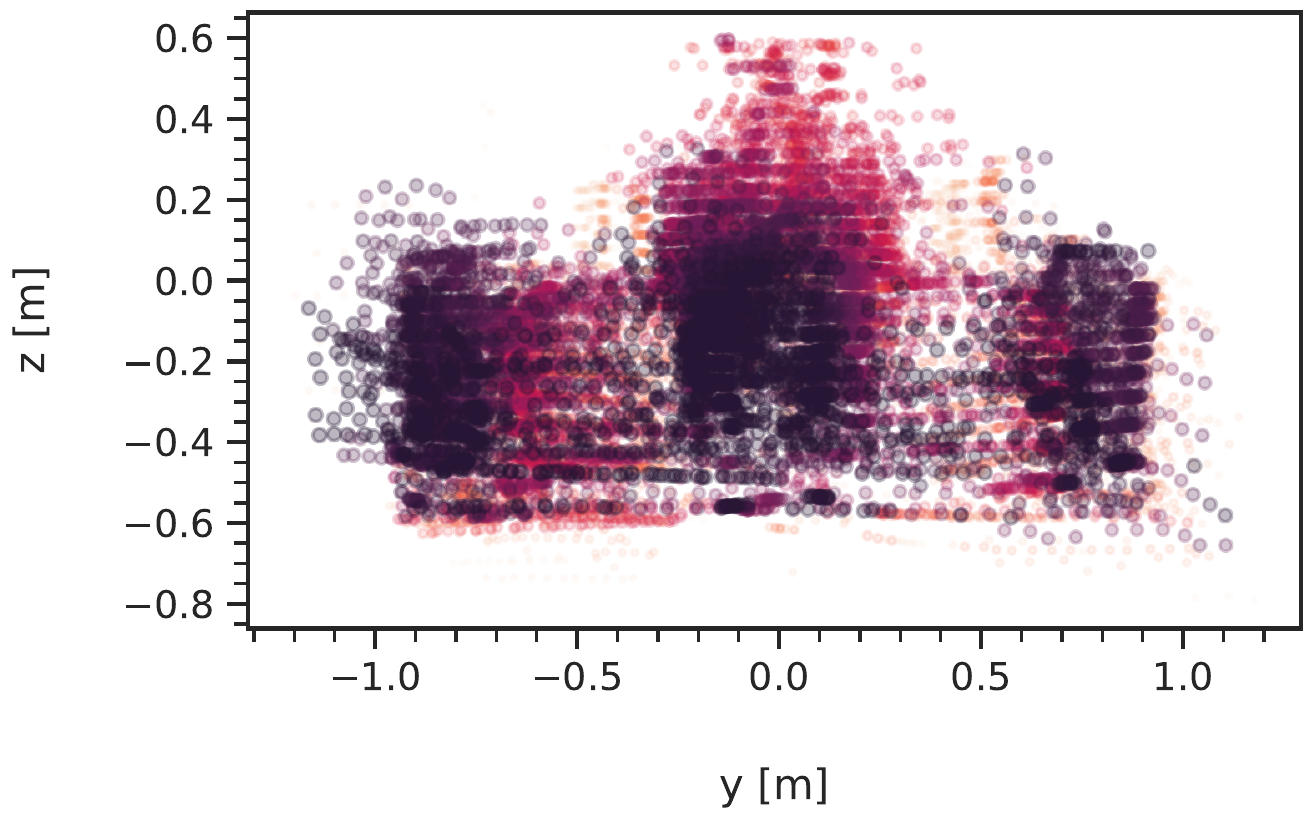}\label{fig:angle_view_front}}}
	\hfill
	\subfloat[\centering Back view]{{\includegraphics[width=0.48\linewidth]{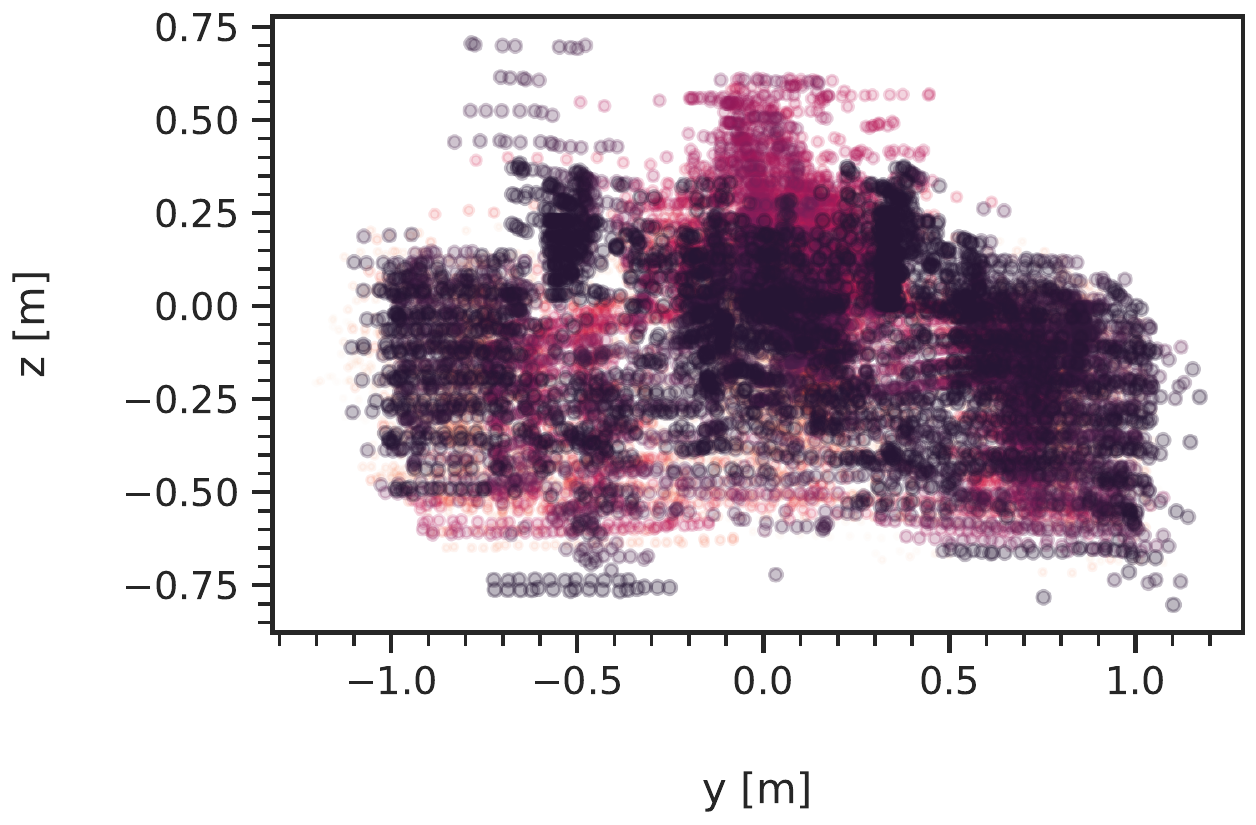}\label{fig:angle_view_back}}}

	\caption{The plots represent slightly more than $20{,}000$ points aggregated from randomly selected scenes of the real dataset. Closer points are darker and less transparent, and cover points are further away.}\label{fig:angle_view}
\end{figure}

Fig.~\ref{fig:angle_view} shows approximately $20{,}000$ aggregated points from $40$ scans of the \textit{real} dataset.
Using a selection of scans and projecting them onto a 2D plane helps to visualize the LiDAR scan layers and the resulting normalized, distorted shape.
These graphs highlight inaccuracies in the automatic point cloud labeling pipeline utilized to create the \textit{real} dataset.
The top view in \ref{fig:angle_view_top} clearly shows that even though the wheels make up a major part of the front view \ref{fig:angle_view_front} and back view \ref{fig:angle_view_back}, the point density in the area of the wheels is sparse.
An explanation for this is the low reflectivity of dark tires and low ray incidence angles on the upper and lower parts of the wheels, both of which lead to points being dropped.
This point dropout is a common occurrence with real-world LiDAR sensors and, if not modeled in simulation, can further increase the sim-to-real domain shift.

To conclude the point cloud target-level analysis, the two discussed effects of LiDAR noise and dropout might have a high impact on the sim-to-real domain shift and will be further analyzed in the following section.

\subsection{Additional Study of Main Influencing Factors}
\label{sec:4d}
Based on previous findings on LiDAR noise and dropout, we generate two new simulated datasets to further quantify the impact of these effects and to analyze the source of the sim-to-real domain shift.
Both datasets originate from the \textit{sim} dataset used in previous experiments.
The first dataset is created to analyze the impact of sensor noise, which was identified to differentiate between the \textit{real} and \textit{sim} point clouds when comparing them at the target-level, as shown in Sec.~\ref{sec:results_target_analysis}.
To create this dataset, we add a noise profile to our sensor simulation that adds random Gaussian noise to the placement of the points in the longitudinal ray direction with a standard deviation of $\sigma~=~\SI{2}{\centi\meter}$.
We refer to this dataset as \textit{sim noise}.
The second dataset is created to analyze the impact of the LiDAR dropout, which was also identified as a domain shift source.
Therefore, we applied downsampling on the original \textit{sim} dataset with a ratio of 0.8, meaning that \SI{20}{\percent} of the points in each point cloud are dropped.
This downsampled dataset is referred to as \textit{sim downsampled}.

\begin{figure*}[t!]
	\centering
        \fontsize{7pt}{6pt}
        \selectfont\def\svgwidth{2.0\columnwidth}
        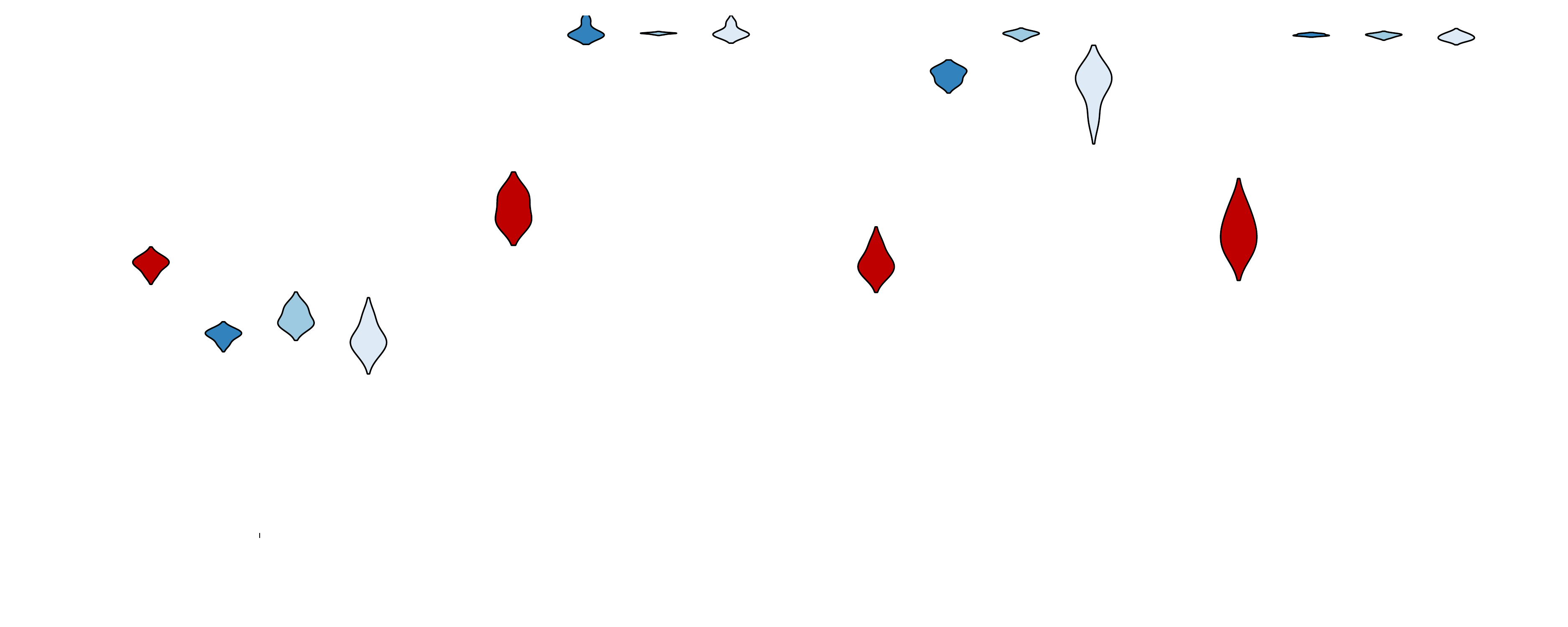
	\caption{Average precision with IoU \SI{70}{\percent} for PointRCNN trained and tested with \textit{real}, \textit{sim}, \textit{sim noise}, or \textit{sim downsampled} data}
	\label{fig:PointRCNN_AP07_all}
\end{figure*}

We train PointRCNN and PointPillars on both derived datasets and evaluate them on all four datasets, that is, \textit{real}, \textit{sim}, \textit{sim noise}, and \textit{sim downsampled}.
The 3D AP (0.7) of PointRCNN for the 16 train-test combinations of the four datasets is shown in Fig.~\ref{fig:PointRCNN_AP07_all}.
Similar to the results in Sec.~\ref{sec:results_obj_det}, the results of PointRCNN and PointPillars are consistent, although the overall AP of PointPillars is lower.
For reference, Table~\ref{tab:all_results} in Appendix \ref{appendix:results} includes the extensive results of both networks for all train-test pairs.
Compared to training with the \textit{sim} dataset, training with \textit{sim noise} is advantageous when testing on the \textit{real} dataset, leading to an AP increase from \SI{38.23}{\percent} (\textit{sim}) to \SI{41.61}{\percent} (\textit{sim noise}) and therefore closing the sim-to-real domain shift.
It can be noted that training with \textit{sim downsampled} compared to \textit{sim} even degrades the performance minimally on all tested datasets.

\begin{figure}[t!]
	\centering
	\subfloat{\includegraphics[width=0.9\columnwidth]{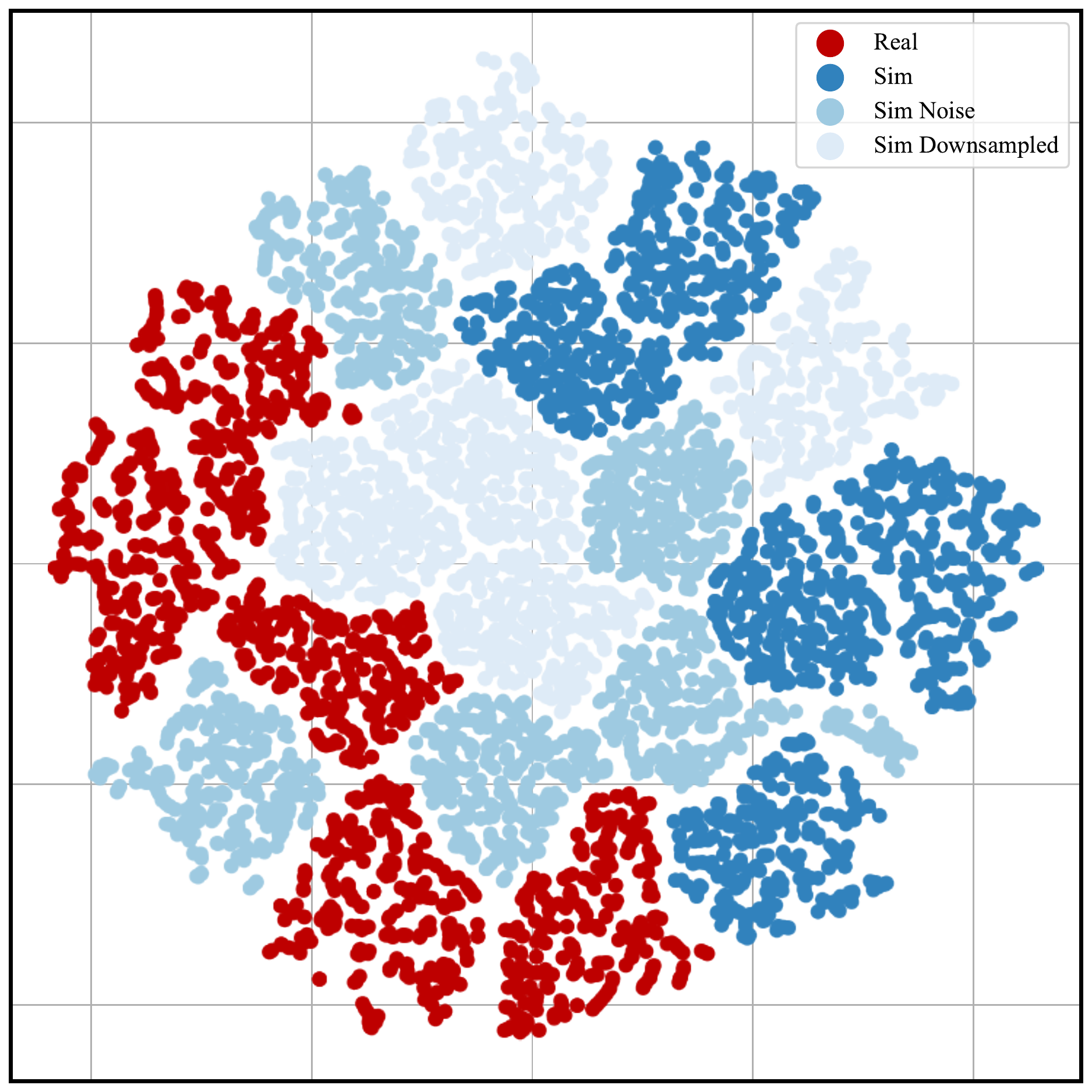}}
	\caption{T-SNE of the latent feature space of PointRCNN trained on \textit{real}, \textit{sim}, \textit{sim noise}, or \textit{sim downsampled} data.}\label{fig:tsne4}%
\end{figure}

Fig.~\ref{fig:tsne4} shows the t-SNE plot for PointRCNN trained on all four datasets and tested on real data.
As in Fig.~\ref{fig:tsne_real_sim}, each trained dataset was tested with $2{,}500$ real point clouds, resulting in $10{,}000$ feature vectors shown in this plot.
PointRCNN can distinguish between the datasets, visible by each of the five training runs per dataset clearly separated from the clusters of the other datasets.
This plot also shows that the clusters of \textit{sim noise} are blended into the clusters of \textit{real}, meaning that there is a greater similarity between \textit{sim noise} and \textit{real} than between \textit{sim} and \textit{real}, which supports the quantitative results.

\begin{figure*}[t!]
	\centering

        \captionsetup{format=hang}
	\subfloat[{Sim Noise,\\ Close-Range $\left[\SI{0.0}{\metre}, \SI{33.3}{\metre}\right[$}\label{fig:car_pcd_sim_noise_0}]{\includegraphics[width=0.33\linewidth]{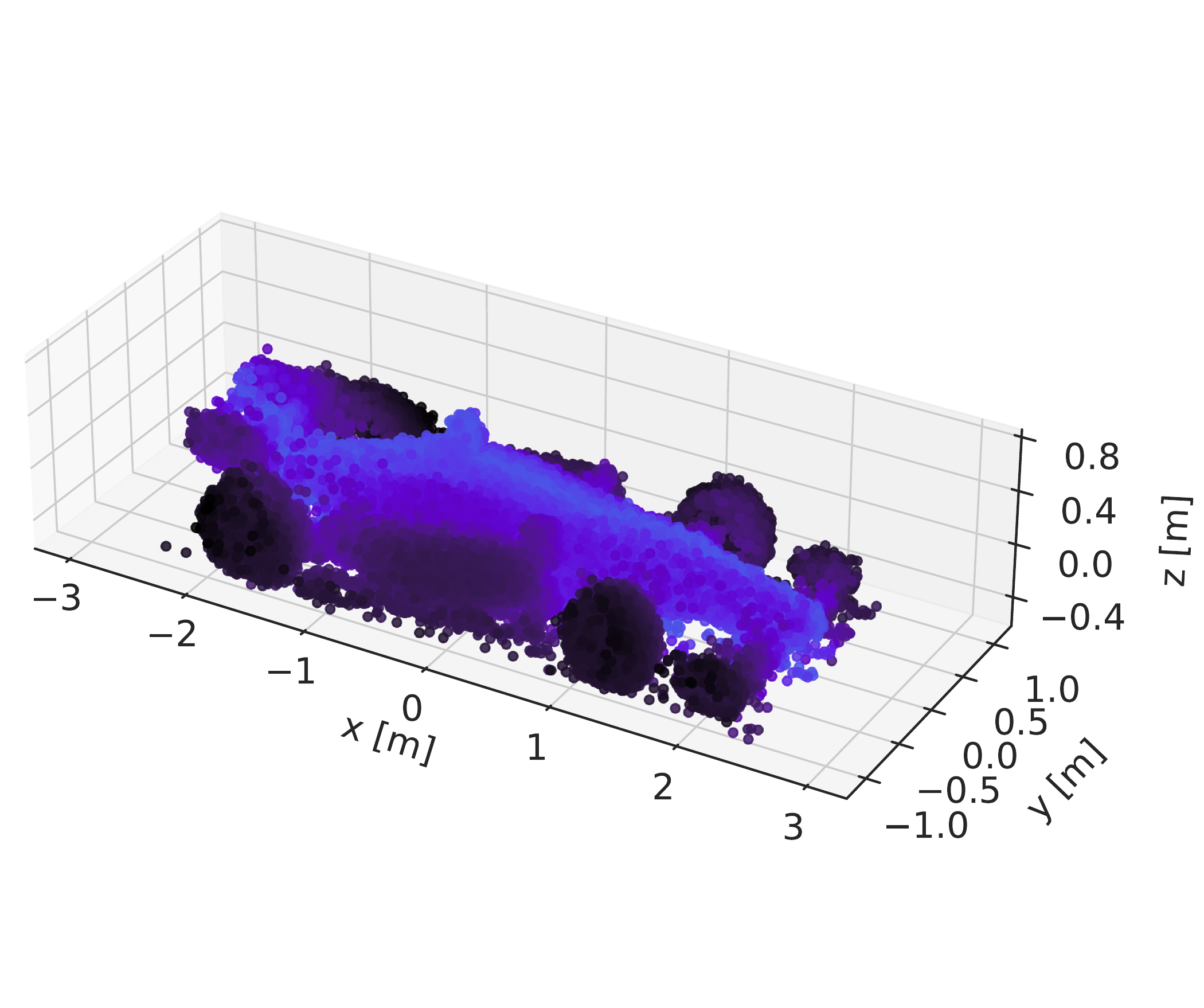}}
	\hfill
	\subfloat[{Sim Noise,\\ Mid-Range $\left[\SI{33.3}{\metre}, \SI{66.6}{\metre}\right[$}\label{fig:car_pcd_sim_noise_33}]{\includegraphics[width=0.33\linewidth]{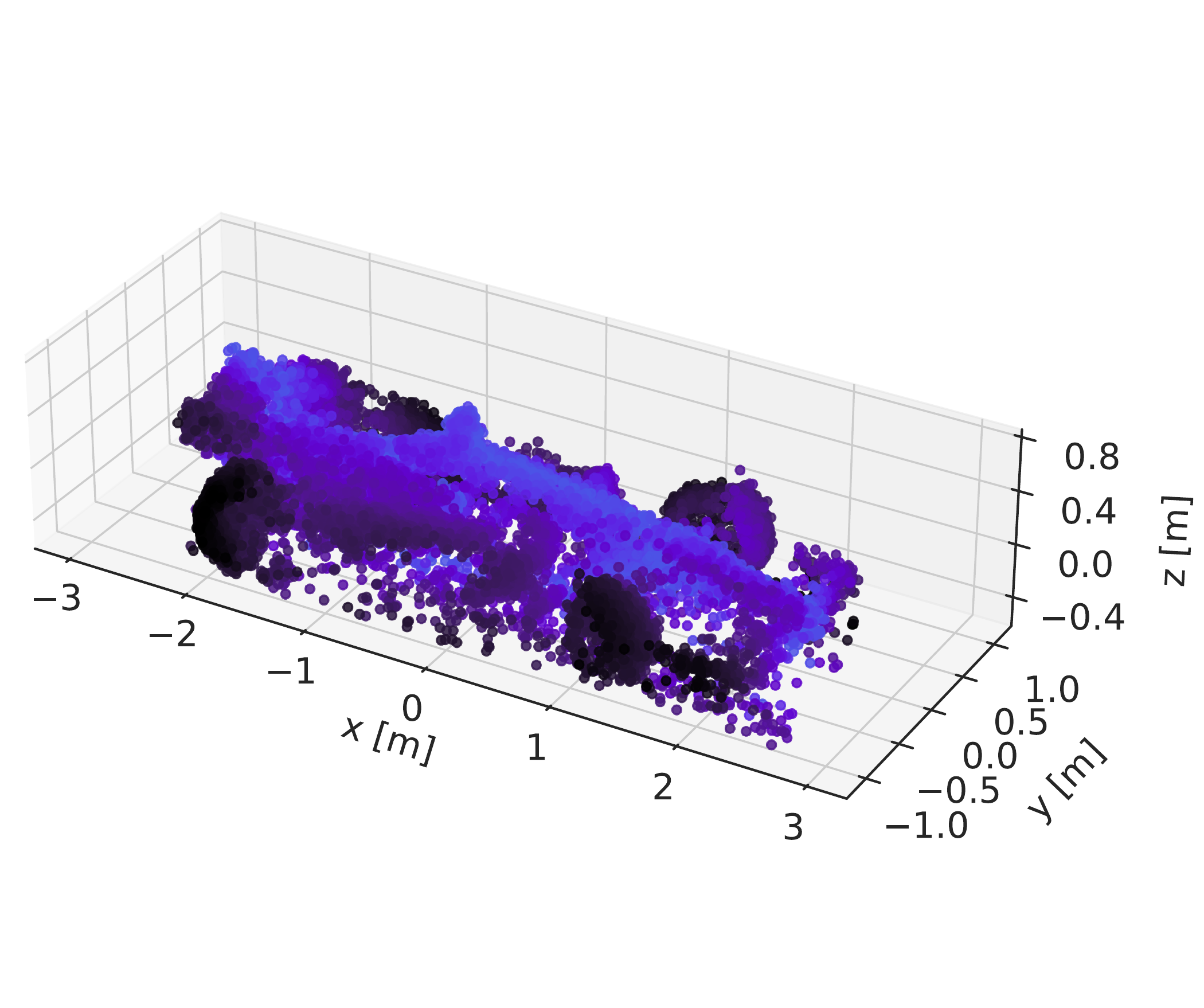}}
	\hfill
	\subfloat[{Sim Noise,\\ Long-Range $\left[\SI{66.6}{\metre}, \SI{100.0}{\metre}\right]$}\label{fig:car_pcd_sim_noise_66}]{\includegraphics[width=0.33\linewidth]{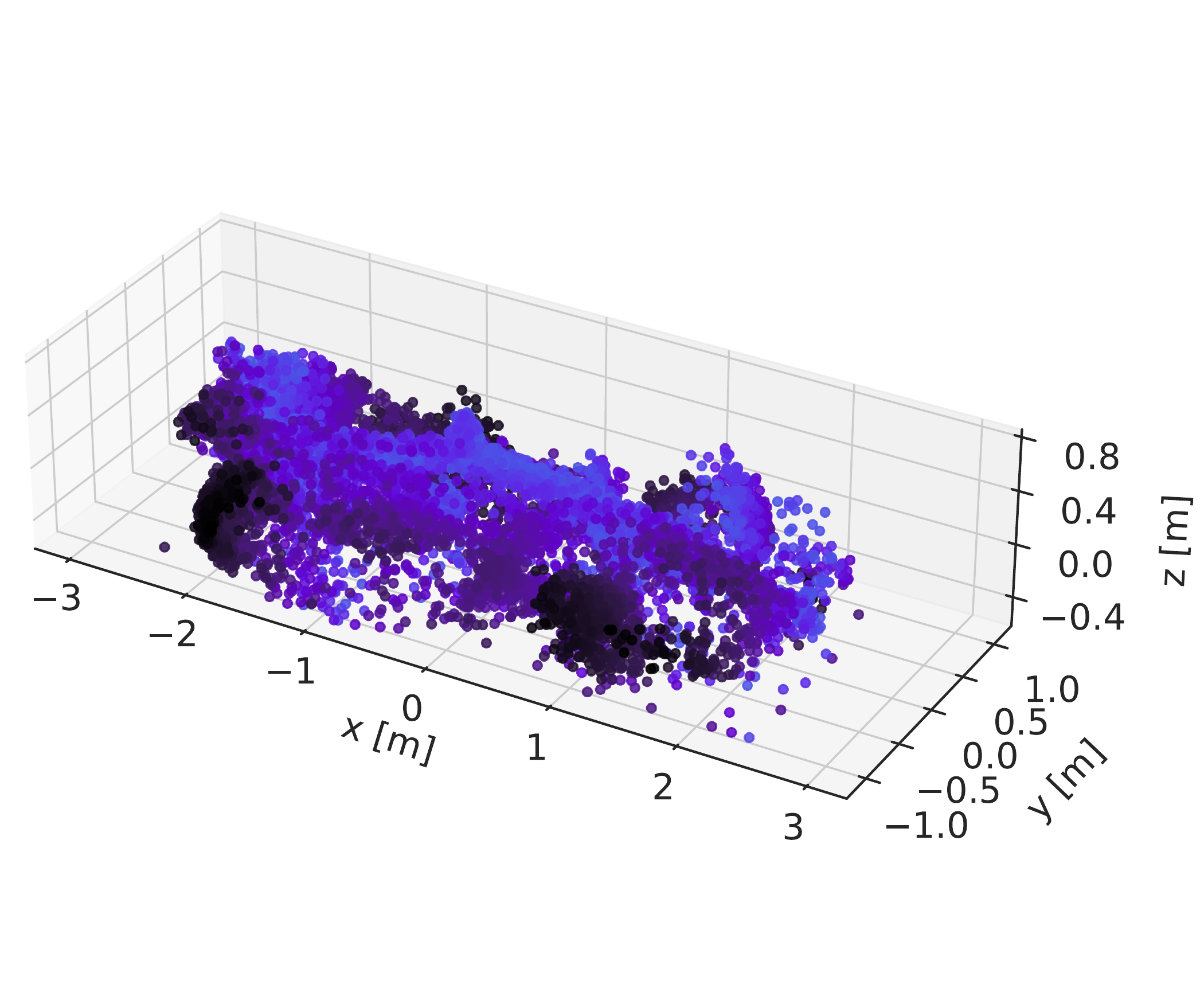}}

	\subfloat[{Sim Downsampled,\\ Close-Range $\left[\SI{0.0}{\metre}, \SI{33.3}{\metre}\right[$}\label{fig:car_pcd_sim_down_0}]{\includegraphics[width=0.33\linewidth]{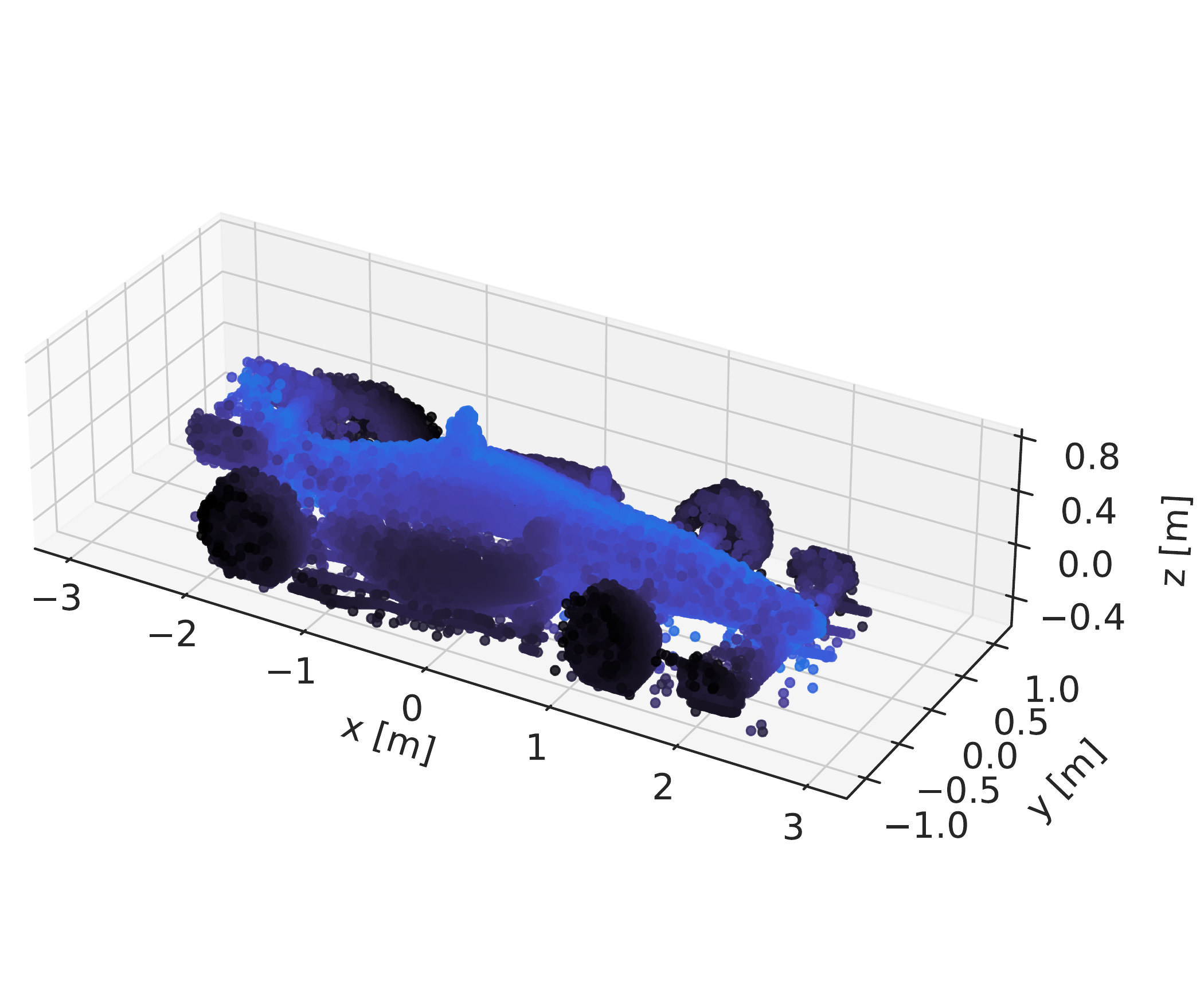}}
	\hfill
	\subfloat[{Sim Downsampled,\\ Mid-Range $\left[\SI{33.3}{\metre}, \SI{66.6}{\metre}\right[$}\label{fig:car_pcd_sim_down_33}]{\includegraphics[width=0.33\linewidth]{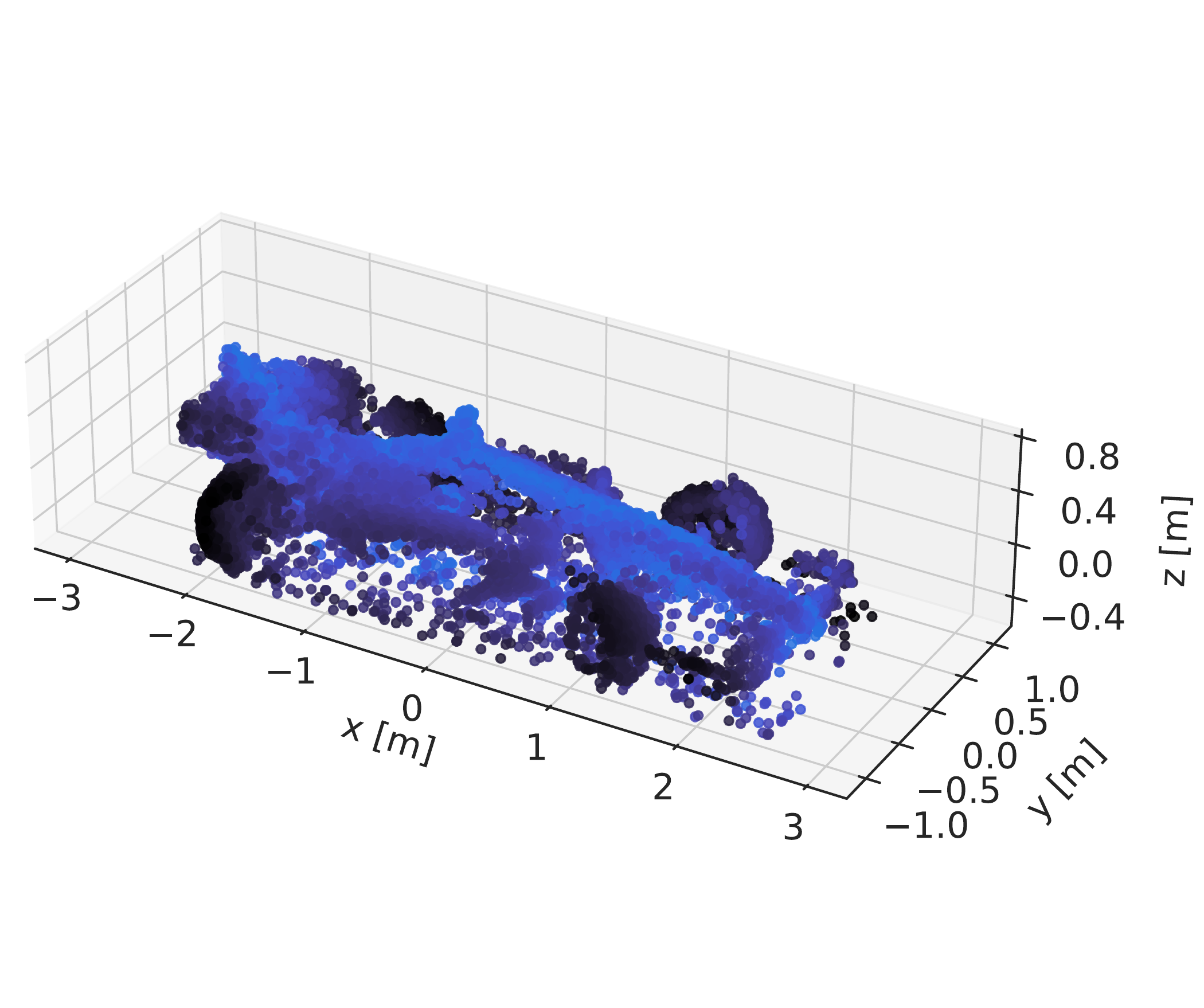}}
	\hfill
	\subfloat[{Sim Downsampled,\\ Long-Range $\left[\SI{66.6}{\metre}, \SI{100.0}{\metre}\right]$}\label{fig:car_pcd_sim_down_66}]{\includegraphics[width=0.33\linewidth]{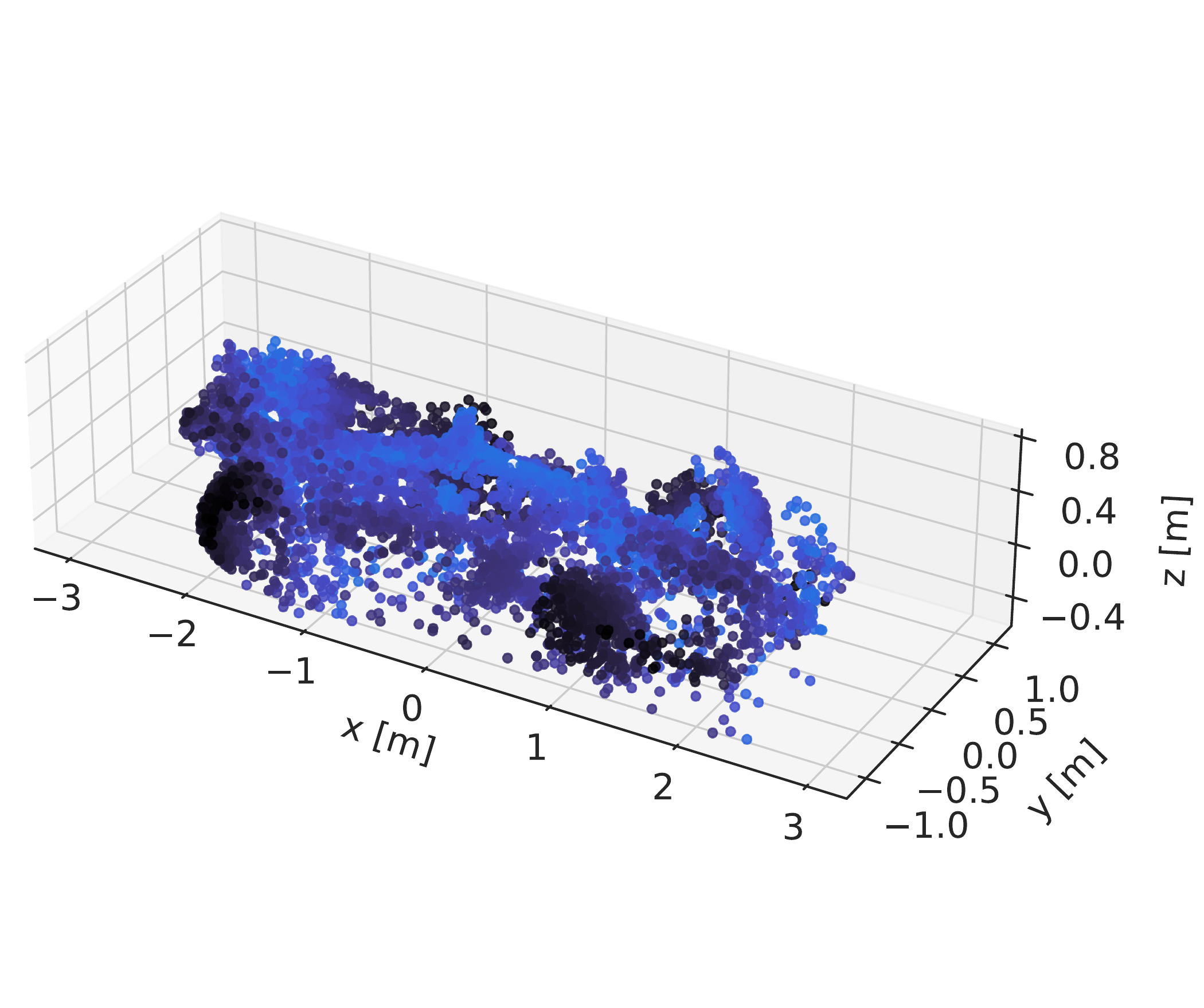}}

	\caption{Aggregated normalized point clouds from the sim noise dataset (first row) and sim downsampled dataset (second row). The columns represent the distance from the target before normalization: close-range \SI{0.0}{\meter} to \SI{33.3}{\meter} (a)(d), mid-range \SI{33.3}{\meter} to \SI{66.6}{\meter} (b)(e), and long-range \SI{66.6}{\meter} to \SI{100}{\meter} (c)(f). We randomly select $20{,}000$ points from all measured target points for close- and mid-range and $9{,}732$ for long-range (all available points of targets in the real dataset at long-range) for visualization purposes.}\label{fig:car_pcd_overview_sim}
\end{figure*}

Fig.~\ref{fig:car_pcd_overview_sim} depicts the aggregated target point clouds of the two additional datasets, \textit{sim noise} and \textit{sim downsampled}, for \mbox{close-,} \mbox{mid-,} and long-range.
The Gaussian noise of the \textit{sim noise} data resembles the \textit{real} data more compared to the original \textit{sim} data.
However, at long-range in Fig.~\ref{fig:car_pcd_sim_noise_66}, the shape of the aggregated target point cloud of \textit{sim noise} is still identifiable as a vehicle, which is not the case for the \textit{real} data in Fig.~\ref{fig:car_pcd_real_66}.
The aggregated target point clouds of \textit{sim downsampled} in Figs. \ref{fig:car_pcd_sim_down_0}-\ref{fig:car_pcd_sim_down_66} are identical to those of the original \textit{sim} data, with the only difference being that \textit{sim downsampled} include \SI{20}{\percent} fewer points.

To compare our results with different metrics for domain shift quantification, we calculate the Chamfer Distance (CD) and Earth Mover's Distance (EMD), which are common metrics for measuring similarity between two point sets.
We calculate these distances for each of the 4,000 point cloud pairings from the training datasets; specifically, we compare the \textit{real} dataset with each of the three sim datasets.
The mean CD and mean EMD for each of the 4,000 point cloud pairings are presented in Table~\ref{tab:cd_emd}.
EMD coincides with the results of the domain shift quantification by means of object detection algorithms, showing a decrease of the sim-to-real domain shift when introducing noise and an increase of the sim-to-real domain shift when adding downsampling.
However, the CD also shows a decrease of the sim-to-real domain shift for \textit{sim downsampled}.
This can be explained by the fact that CD is highly sensitive to outliers, which are less frequent in \textit{sim downsampled} and therefore lead to a lower CD.

\begin{table}[]
\centering
\caption{Chamfer Distance (CD) and Earth Mover's Distance (EMD) calculated for every point cloud pairing between domain A and domain B. Note that the values represent the mean CD and mean EMD of all point cloud pairings for the entire training datasets. Lower is better.}
\label{tab:cd_emd}
\begin{tabular}{@{}llll@{}}
\toprule
Domain A              & Domain B        & CD        & EMD    \\ \midrule
\multirow{3}{*}{Real} & Sim             & 2,333,682 & 17.322 \\
                      & Sim Noise       & 2,328,009 & 17.309 \\
                      & Sim Downsampled & 2,288,593 & 17.324 \\ \bottomrule
\end{tabular}
\end{table}

\section{Discussion}
\label{sec:discussion}

In this section, we discuss the results presented in Sec \ref{sec:results}.
The statistical dataset comparison in Sec.~\ref{sec:results_stat_data_comp} shows that although the \textit{sim} dataset is derived from the \textit{real} dataset, there are still differences between the two datasets.
For example, the 3D simulation environment shows deviations from reality outside of the drivable area due to missing static objects.
This can be observed in Fig.~\ref{fig:full_pcd_bev} with more points in the real point cloud behind the track wall.
However, we show that the distributions of points and targets between the datasets \textit{sim} and \textit{real} are very similar, which makes our datasets suitable for domain shift analysis.

Our evaluation based on object detection algorithms reveals the existence of a sim-to-real domain shift, proving that the domain shift found in previous works is not only based on scenario discrepancy due to dissimilarities in the datasets used but is also present in our scenario-identical \textit{real} and \textit{sim} datasets.
As expected, the networks achieve the highest performance when trained and tested on the same domain (real-to-real and sim-to-sim).
However, for PointRCNN trained on \textit{real} data, testing on \textit{sim} data (real-to-sim) yields a higher AP than testing on the domain with which it was trained (real-to-real), which is counterintuitive at first.
This performance increase can be explained by analyzing the target point clouds in Fig.~\ref{fig:car_pcd_overview}: Target point clouds in \textit{sim} data are more structured with less noise and denser than the same point clouds in \textit{real} data, especially at higher ranges, leading to an improved detection performance regardless of the type of dataset used for training.

The GPS data and point cloud data recorded with the real-world vehicle are not synchronized, leading to minor inaccuracies in auto-labeling due to the data interpolation of discrete-time stamps.
As described in Sec.~\ref{sec:method_dataset_generation}, the target positions are refined using the point distribution in the proximity of the initial position determined by the auto-labeling process.
This leads to an overall good fit of the 3D bounding boxes to the underlying point cloud.
However, in some frames, there is an offset of the 3D bounding box and the target point cloud; this is visible in Fig.~\ref{fig:angle_view_side}, with one frame shifted to the vehicle's rear by about \SI{0.8}{\meter}.
Those labeling inaccuracies are more frequent at long range due to lower point density and self-occlusion, leading to low average precision of real-to-real at long range compared to close range.
Furthermore, PointRCNN and PointPillars only estimate seven degrees of freedom for each object and neglect the roll and pitch angles.
Our auto-labeling pipeline also defines these two angles to be zero.
However, these two angles are more pronounced at higher distances, leading to an additional labeling offset and further explaining the low real-to-real performance at long range.

Fig.~\ref{fig:car_pcd_overview} shows the aggregated target point clouds in different range sections.
All range sections show noisy measurements for the \textit{real} and \textit{sim} data, although the \textit{sim} data in these graphs are simulated without noise.
Part of the noise beneath the vehicle comes from the labeling process and is not specific to either \textit{real} or \textit{sim} data.
For data labeling, the location and size of the vehicle are used. All points in the vehicle-sized bounding box around the GPS location will be considered part of the vehicle. 
This assumption is valid for most point clouds, except when the inclination of the road leads to ground reflections being considered as part of the target point cloud, which occurs more frequently for longer distances.
In this way, structured noise, such as lines, is introduced.
These noise artifacts can be seen in the lower area between the tires.
It is important to keep in mind that a segmentation mask is not predicted, but a bounding box is instead; thus, such effects are expected.

In this work, we quantify the domain shift based on the task of object detection and not semantic or instance segmentation, which are also very common tasks using LiDAR point clouds.
We focus on object detection, as segmentation requires pointwise labeled point clouds.
These are costly to obtain for real-world data and, compared to the position and orientation of the target boxes, can not be generated by an auto-labeling pipeline.
However, since segmentation offers a more fine-grained understanding of the environment, the analysis of the domain gap using segmentation algorithms can be investigated in follow-up work.

The additional study of noise and downsampling in the simulation on the performance of object detectors demonstrated a performance increase for \textit{sim noise} and a performance decrease for \textit{sim downsampled} compared to the original \textit{sim} data.
Although the performance increase of \textit{sim noise} was expected because it more closely modeled the behavior in the real world, the performance decrease of \textit{sim downsampled} needs further explanation.
The downsampling method used in this work is based on random selection and is not based on physical aspects, such as the angle of inclination or the material and color of the reflected surface, as these properties are not available in the simulation environment and increase the simulation complexity to a high degree.
Therefore, the utilized downsampling method is unfavorable to object detection performance.

To eliminate the downsampling method as a source of the increase in domain shift, we experimented with downsampling that is not random, but based on the point distribution difference between the \textit{real} and \textit{sim} dataset.
Each point within a point cloud is assigned with a probability of being removed, which is derived from the previously calculated distribution difference of \textit{real} and \textit{sim} datasets.
This leads to the alignment of point distributions of the \textit{real} and \textit{sim downsampled} datasets; specifically, long-range points are more likely to be downsampled, as the \textit{real} dataset is more sparse at higher distances.
Although the resulting \textit{sim downsampled} dataset reduces the difference in the distribution of points, the performance of PointRCNN and PointPillars in terms of 3D AP for all evaluated datasets is almost identical to the random downsampled dataset and the differences are within the standard deviation.
The influence of a more realistic simulation regarding the downsampling effect based on physical aspects can be investigated in future work.

Visualization of the high-dimensional feature vectors of the network using t-SNE dimensionality reduction revealed that the network could distinguish between \textit{real} and \textit{sim} data and that \textit{sim noise} is more similar to \textit{real} than \textit{sim} is to \textit{real}.
However, the t-SNE plot also implies that \textit{sim downsampled} is more similar to \textit{real} than \textit{sim} is to \textit{real}, which does not coincide with the results of the quantitative evaluation of the networks.
Note that t-SNE does not always preserve global structures and is based primarily on local structures of the input data \cite{VanDerMaaten2008VisualizingT-SNE}.

Chamfer Distance (CD) and Earth Mover's Distance (EMD) were used to compare our results with other metrics used to measure the domain similarity between two point sets.
Although CD indicates a higher domain similarity of \textit{sim downsampled} to \textit{real} than \textit{sim} to \textit{real} due to the sensitivity to outliers, we could validate our findings with the results of EMD.

Finally, there are limitations regarding the datasets we used for the domain shift quantification.
Our requirements were a scenario-identical pair of \textit{real} and \textit{sim} datasets to isolate the domain shift.
We achieved this by using a dataset that is limited to simple scenarios and only one class of objects to be detected.
We still argue that the datasets used in our work give a good indication of the LiDAR domain shift and can generalize well to other datasets since the LiDAR noise and dropout differences we highlighted in our additional study usually exist between \textit{real} and \textit{sim} data, independent of specific scenarios or the number of object classes.
For future work, the domain shift of datasets with more complex scenarios and a higher variety of vehicles can be analyzed using our presented method.
Furthermore, taking the pointwise $intensity$-values into account can be investigated in future work.
\vspace{-5pt}

\section{Conclusion}
\label{sec:conclusion}

To summarize, in this paper, we quantify the sim-to-real domain shift by means of LiDAR object detectors and further analyze the point clouds at the target-level to determine the influence factors leading to domain shift.
First, we record a real-world dataset, which is auto-labeled using GPS positions, and generate a simulated counterpart whose scenarios are derived from the real-world dataset.
We show that the data and label distributions of both datasets are similar and hence, both datasets form the basis of our domain shift analysis.

To perform domain shift quantification, we trained two LiDAR object detection networks, namely PointRCNN and PointPillars, on the \textit{sim} and \textit{real} datasets and tested each network on both datasets.
The evaluation metric is the average precision, and since training is non-deterministic, we train each network five times and report the mean of the achieved average precision.
The experiments show the existence of a domain shift in both directions, whereas the sim-to-real domain shift amounts to around \SI{14}{\percent} in AP difference.
We further analyzed the target point clouds by aggregation of single point clouds for three different range sections, that is, close-, mid-, and long-range.
For long-range \textit{real} data, the shape of the target is not identifiable as a vehicle, which explains the poor network performance for this range section.

The aggregated target point clouds further indicate a qualitative sim-to-real domain shift, and we identify noise and downsampling as two potential factors of domain shift.
We generated additional simulated datasets modeling these two factors to analyze them further.
Object detection experiments with \textit{sim noise} and \textit{sim downsampled} show that noise does indeed have an influence on the sim-to-real domain shift; this is evident from the reduction of the sim-to-real domain shift from \SI{14}{\percent} to \SI{10}{\percent}.
However, the downsampling method used in our work did not reduce the domain shift but even slightly increased it.

Overall, we introduced a method for the quantification of the LiDAR sim-to-real domain shift based on object detectors and quantified the domain shift for distribution-aligned datasets.
Our experiments showed that noise is part of the domain shift; however, there are still other effects that contribute to the domain shift.
The analysis of these effects, as well as a more realistic simulation of downsampling based on physical aspects, is part of future work.
Furthermore, one possibility for future work is the application of domain adaptation for LiDAR point clouds to reduce the sim-to-real domain shift.

\vspace{-5pt}

\section*{Contributions}
As the first author, Sebastian Huch initiated the idea of this paper and contributed essentially to its conception, implementation, and content. Luca Scalerandi and Esteban Rivera contributed to the conception of this research, the experimental data generation, and the revision of the research article. Markus Lienkamp made an essential contribution to the conception of the research project. He revised the paper critically for important intellectual content. He gave final approval of the version to be published and agreed to all aspects of the work. As a guarantor, he accepts responsibility for the overall integrity of the paper.
\vspace{-5pt}
\bibliographystyle{IEEEtran}
\bibliography{IEEEabrv,references}
\vspace{-10pt}

\begin{IEEEbiography}
[{\includegraphics[width=1in,height=1.25in,clip,keepaspectratio]{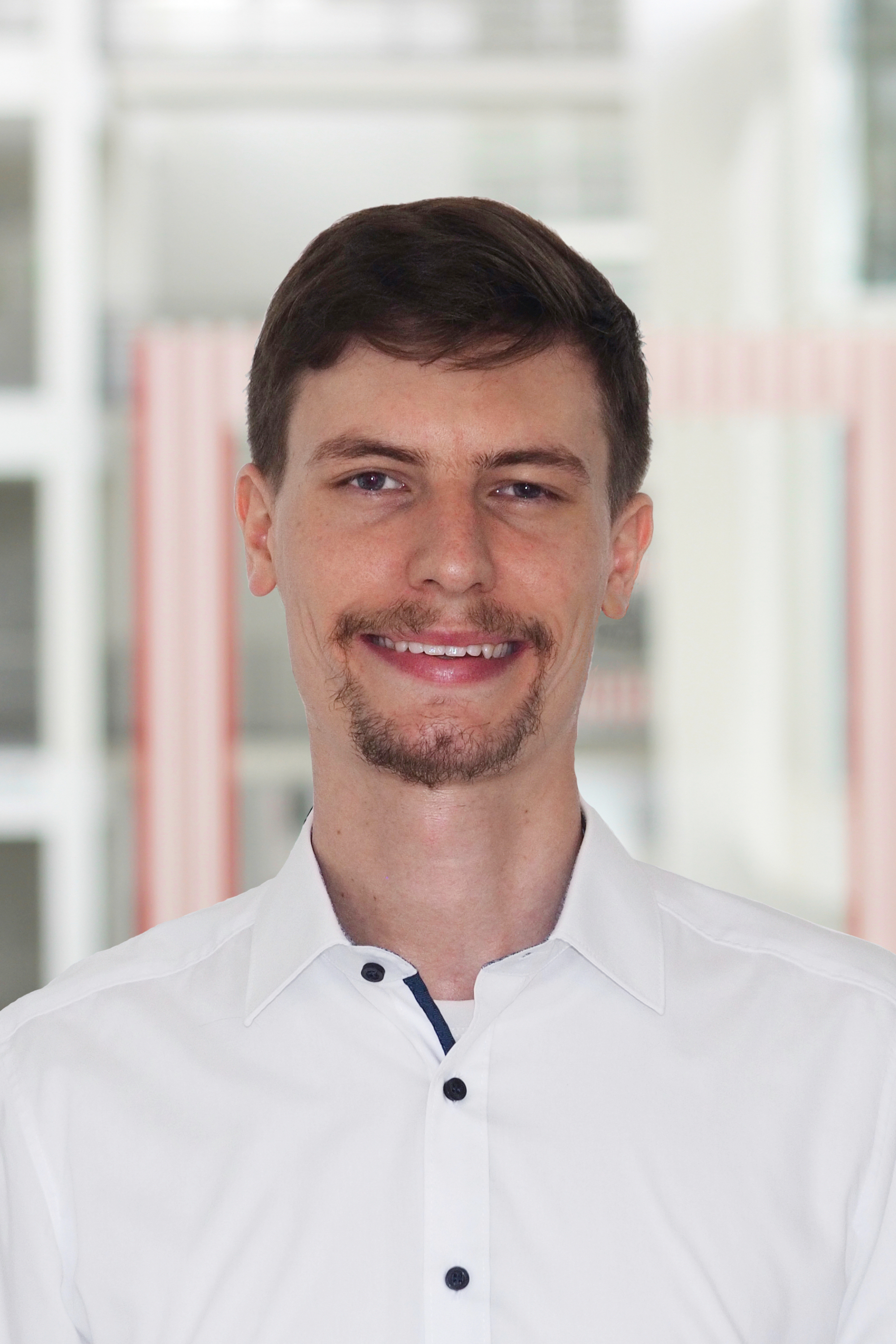}}]{Sebastian Huch}
received his BEng degree from the Baden-Wuerttemberg Cooperative State University (DHBW) Stuttgart, Germany, in 2016 and his  MSc degree from the Technical University of Darmstadt, Germany, in 2018. He is currently pursuing his PhD degree in mechanical engineering at the Institute of Automotive Technology at the Technical University of Munich (TUM), Germany. His research interests include LiDAR simulation, LiDAR perception, and LiDAR domain adaptation for autonomous driving.
\end{IEEEbiography}

\vspace{-30pt}

\begin{IEEEbiography}
[{\includegraphics[width=1in,height=1.25in,clip,keepaspectratio]{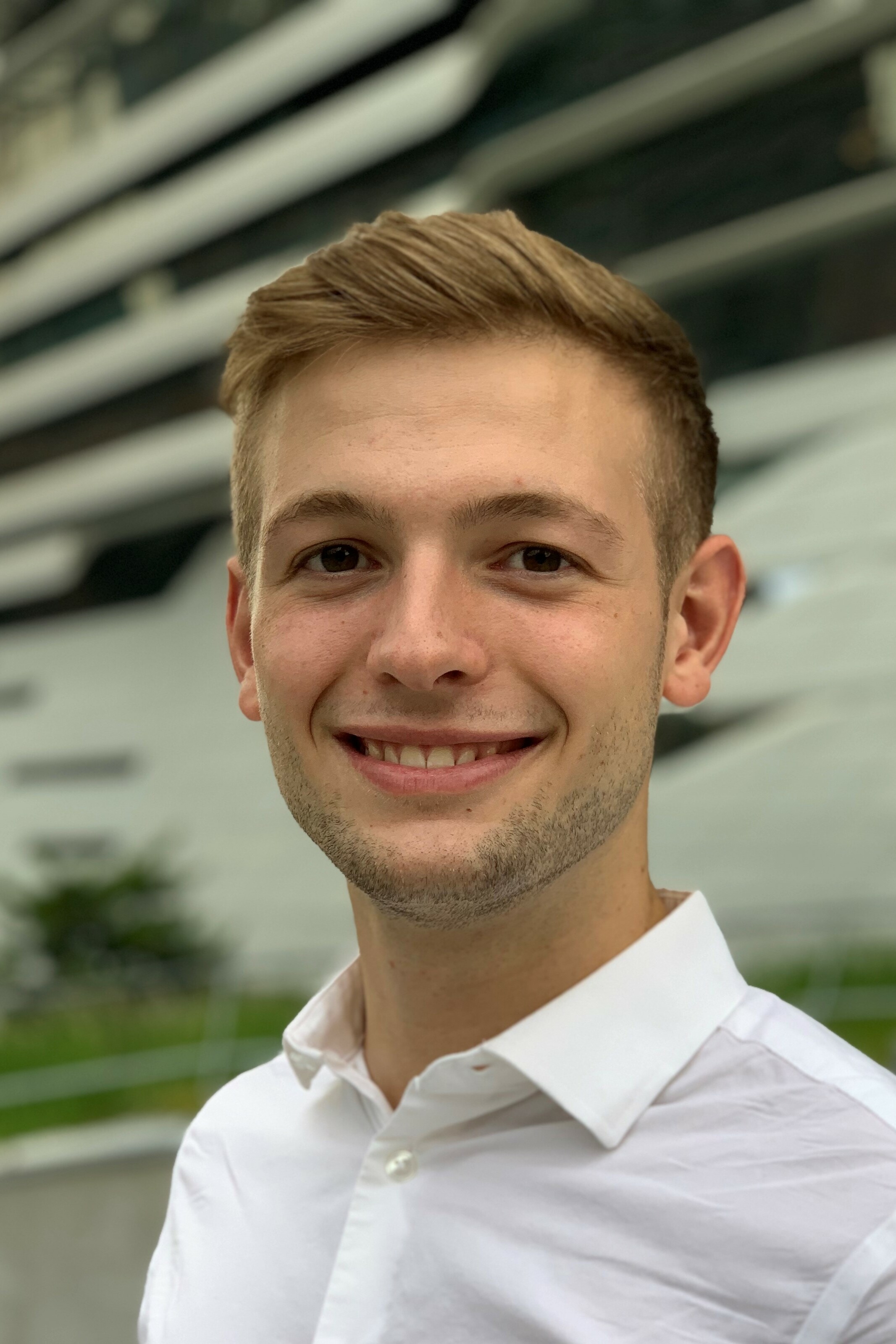}}]{Luca Scalerandi}
received his BSc degree in computer science from the Technical University of Munich (TUM), Munich, Germany, in 2021. He is currently pursuing his master's degree in computer science at TUM. He works part-time as a computer vision engineer at DeepScenario. His broader scientific interest lies in multi-object tracking, motion analysis, and scenario understanding. 
\end{IEEEbiography}

\vspace{-30pt}

\begin{IEEEbiography}
[{\includegraphics[width=1in,height=1.25in,clip,keepaspectratio]{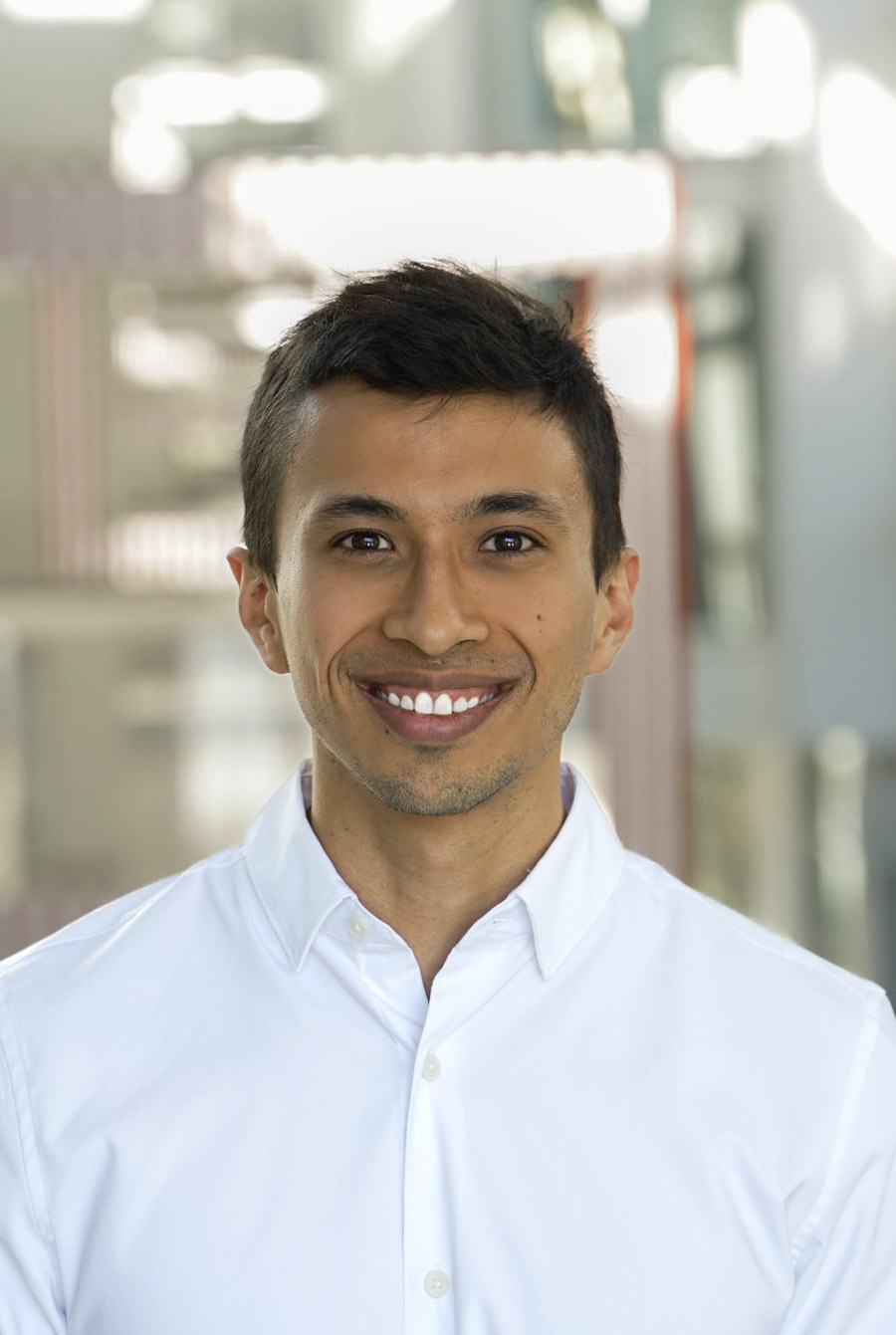}}]{Esteban Rivera}
received his BSc degree in Electronic Engineering and Physics from the Universidad de los Andes, Bogotá. Colombia in 2016 and his MSc in Electrical Engineering from the Karlsruhe Institute for Technology in 2019. Later he worked as a data scientist for Appgate Inc and developed authentication algorithms based on deep learning for the finance industry. Currently, he is pursuing his PhD at the Institute for Automotive Technology at the Technical University of Munich (TUM), Germany. His research interests include computer vision, camera-based object detection, and sensor fusion.  
\end{IEEEbiography}

\vspace{-30pt}

\begin{IEEEbiography}
[{\includegraphics[width=1in,height=1.25in,clip,keepaspectratio]{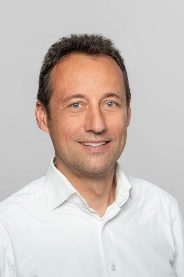}}]{Markus Lienkamp}
carries out research in the area of autonomous vehicles with the objective of creating an open-source software platform. He is a professor at the Institute of Automotive Technology at the Technical University of Munich (TUM) and is involved in the CREATE project in Singapore. After studying mechanical engineering at TU Darmstadt and Cornell University, Prof. Lienkamp obtained his doctorate from TU Darmstadt (1995). He worked at Volkswagen as part of an international trainee program and took part in a joint venture between Ford and Volkswagen in Portugal. Returning to Germany, he led the brake testing department of the VW commercial vehicle development section in Wolfsburg. He was later appointed head of the Electronics and Vehicle research department in the Volkswagen Group's Research Division. His main priorities were advanced driver assistance systems and electromobility concepts. Prof. Lienkamp has headed the Chair of Automotive Technology at TUM since November 2009.
\end{IEEEbiography}
\flushcolsend 

\onecolumn
\appendix
\label{appendix:results}

\begin{table*}[ht!]
\centering
\caption{Average Precision (AP) and Recall for the two object detection networks trained and tested on four different datasets. Both AP and Recall are evaluated for different Intersection-over-Union thresholds, \SI{50}{\percent} and \SI{70}{\percent}, denoted as 3D~AP~(0.5) and 3D~AP~(0.7), respectively. Values represent the mean of five identical training runs with the standard deviation in brackets. The test is carried out using the entire point cloud range that includes points up to a radial distance of \SI{100}{m}. All values in percent.}
\label{tab:all_results}
\begin{tabular}{@{}lllllll@{}}
\toprule
Network & Train dataset                   & Test dataset   & 3D AP (0.5)  & 3D AP (0.7)  & Recall (0.5) & Recall (0.7) \\ \midrule
\multirow{16}{*}{PointRCNN}    & \multirow{4}{*}{Real}            & Real & 74.33 (1.9)  & 51.96 (1.21) & 92.12 (0.7)  & 80.92 (1.21) \\
        &                                  & Sim             & 86.56 (1.0)  & 62.53 (2.62) & 93.78 (0.48) & 87.26 (0.88) \\
        &                                  & Sim Noise       & 77.27 (2.08) & 52.25 (2.15) & 87.88 (0.74) & 77.48 (0.98) \\
        &                                  & Sim Downsampled & 83.64 (0.72) & 58.17 (3.37) & 92.7 (0.5)   & 84.62 (0.89) \\ \cmidrule(l){2-7} 
        & \multirow{4}{*}{Sim}             & Real            & 56.29 (1.36) & 38.23 (0.98) & 76.56 (0.78) & 65.5 (0.95)  \\
        &                                  & Sim             & 97.31 (1.35) & 96.82 (1.16) & 99.46 (0.1)  & 98.64 (0.21) \\
        &                                  & Sim Noise       & 88.96 (1.11) & 88.48 (1.19) & 94.4 (0.82)  & 92.96 (0.83) \\
        &                                  & Sim Downsampled & 96.31 (0.18) & 96.28 (0.17) & 99.04 (0.14) & 98.08 (0.26) \\ \cmidrule(l){2-7} 
        & \multirow{4}{*}{Sim Noise}       & Real            & 60.36 (1.35) & 41.61 (1.65) & 78.3 (0.63)  & 67.8 (0.78)  \\
        &                                  & Sim             & 97.07 (1.07) & 96.58 (0.12) & 99.42 (0.12) & 98.48 (0.16) \\
        &                                  & Sim Noise       & 96.93 (1.29) & 96.43 (0.43) & 99.08 (0.13) & 98.16 (0.21) \\
        &                                  & Sim Downsampled & 96.21 (0.29) & 96.19 (0.29) & 99.08 (0.25) & 97.92 (0.3)  \\ \cmidrule(l){2-7} 
                               & \multirow{4}{*}{Sim Downsampled} & Real & 55.84 (2.72) & 37.57 (2.44) & 75.98 (0.55) & 65.72 (0.17) \\
        &                                  & Sim             & 97.33 (1.08) & 96.82 (0.94) & 99.3 (0.11)  & 98.48 (0.32) \\
        &                                  & Sim Noise       & 86.86 (2.48) & 86.38 (3.31) & 92.78 (1.91) & 90.74 (2.29) \\
        &                                  & Sim Downsampled & 95.93 (0.5)  & 95.87 (0.54) & 99.02 (0.13) & 98.12 (0.19) \\ \midrule
\multirow{16}{*}{PointPillars} & \multirow{4}{*}{Real}            & Real & 69.22 (0.0)  & 41.1 (0.0)   & 80.5 (0.0)   & 55.7 (0.0)   \\
        &                                  & Sim             & 68.65 (0.03) & 20.68 (0.0)  & 83.1 (0.0)   & 41.0 (0.0)   \\
        &                                  & Sim Noise       & 67.32 (0.0)  & 19.98 (0.0)  & 83.2 (0.0)   & 40.7 (0.0)   \\
        &                                  & Sim Downsampled & 63.44 (0.0)  & 18.41 (0.02) & 82.3 (0.0)   & 39.7 (0.0)   \\ \cmidrule(l){2-7} 
        & \multirow{4}{*}{Sim}             & Real            & 30.49 (0.02) & 13.36 (0.12) & 69.7 (0.0)   & 39.9 (0.0)   \\
        &                                  & Sim             & 98.75 (0.0)  & 98.18 (0.0)  & 99.7 (0.0)   & 98.9 (0.0)   \\
        &                                  & Sim Noise       & 98.98 (0.0)  & 98.63 (0.0)  & 99.8 (0.0)   & 99.1 (0.0)   \\
        &                                  & Sim Downsampled & 98.63 (0.0)  & 98.11 (0.0)  & 99.6 (0.0)   & 98.7 (0.0)   \\ \cmidrule(l){2-7} 
        & \multirow{4}{*}{Sim Noise}       & Real            & 39.81 (0.02) & 18.77 (0.08) & 68.8 (0.0)   & 43.2 (0.0)   \\
        &                                  & Sim             & 99.41 (0.0)  & 99.22 (0.0)  & 99.8 (0.0)   & 98.7 (0.0)   \\
        &                                  & Sim Noise       & 98.95 (0.0)  & 98.51 (0.0)  & 99.7 (0.0)   & 98.3 (0.0)   \\
        &                                  & Sim Downsampled & 99.07 (0.0)  & 98.71 (0.0)  & 99.8 (0.0)   & 98.0 (0.0)   \\ \cmidrule(l){2-7} 
        & \multirow{4}{*}{Sim Downsampled} & Real            & 30.25 (0.16) & 14.05 (0.02) & 67.0 (0.0)   & 39.1 (0.0)   \\
        &                                  & Sim             & 99.15 (0.0)  & 94.96 (0.0)  & 99.6 (0.0)   & 96.7 (0.0)   \\
        &                                  & Sim Noise       & 98.73 (0.0)  & 95.28 (0.0)  & 99.5 (0.0)   & 96.9 (0.0)   \\
        &                                  & Sim Downsampled & 98.31 (0.0)  & 94.46 (0.0)  & 99.6 (0.0)   & 97.5 (0.0)   \\ \bottomrule
\end{tabular}
\end{table*}

\end{document}